\def\year{2022}\relax
\documentclass[letterpaper]{article} 
\usepackage{aaai22}  
\usepackage{times}  
\usepackage{helvet}  
\usepackage{courier}  
\usepackage[hyphens]{url}  
\usepackage{graphicx} 
\usepackage{natbib}  
\usepackage{caption} 
\DeclareCaptionStyle{ruled}{labelfont=normalfont,labelsep=colon,strut=off} 
\usepackage{algorithm}
\usepackage{algorithmic}
\usepackage{arydshln}
\usepackage{bm}
\usepackage{multirow}
\usepackage{wrapfig}
\usepackage[dvipsnames, svgnames, x11names]{xcolor}
\usepackage{enumitem}
\usepackage{amssymb}
\usepackage{amsmath}
\usepackage{mathrsfs}
\usepackage{bibentry}

\usepackage{newfloat}
\usepackage{listings}
\floatstyle{ruled}
\floatname{listing}{Listing}
\usepackage{bbding}
 
\usepackage[switch]{lineno}

\newcommand{\ieno}{\textit{i}.\textit{e}.} 
\newcommand{\eg}{\textit{e}.\textit{g}.~}
\newcommand{\egno}{\textit{e}.\textit{g}.} 

\newcommand{\etcno}{\textit{etc}} 

\title{SelectAugment: Hierarchical Deterministic Sample Selection \\ for Data Augmentation }
\author {
    Shiqi Lin\textsuperscript{1}\thanks{Equal contribution. } \quad
    Zhizheng Zhang\textsuperscript{2}\textsuperscript{*} \quad
    Xin Li\textsuperscript{1} \quad
    Wenjun Zeng\textsuperscript{2} \quad
    Zhibo Chen\textsuperscript{1}\thanks{Corresponding Author.} \\
    {\normalsize University of Science and Technology of China}\textsuperscript{1} \quad
    {\normalsize Microsoft Research Asia}\textsuperscript{2} \\
    {\vspace{-0.2em}\tt \small \{linsq047,lixin666\}@mail.ustc.edu.cn \; \{zhizzhang,wezeng\}@microsoft.com \; chenzhibo@ustc.edu.cn} \\
}

\usepackage{bibentry}

\begin{document}

\maketitle

\begin{abstract}

Data augmentation (DA) has been widely investigated to facilitate model optimization in many tasks. However, in most cases, data augmentation is randomly performed for each training sample with a certain probability, which might incur content destruction and visual ambiguities. To eliminate this, in this paper, we propose an effective approach, dubbed SelectAugment, to select samples to be augmented in a deterministic and online manner based on the sample contents and the network training status. Specifically, in each batch, we first determine the augmentation ratio, and then decide whether to augment each training sample under this ratio. We model this process as a two-step Markov decision process and adopt Hierarchical Reinforcement Learning (HRL) to learn the augmentation policy. In this way, the negative effects of the randomness in selecting samples to augment can be effectively alleviated and the effectiveness of DA is improved. Extensive experiments demonstrate that our proposed SelectAugment can be adapted upon numerous commonly used DA methods, \egno, Mixup, Cutmix, AutoAugment, \etcno, and improve their performance on multiple benchmark datasets of image classification and fine-grained image recognition.
\end{abstract}



\section{Introduction}

Data augmentation (DA) is an effective technique to foster model optimization by improving the amount and diversity of training data, which has been widely used in various tasks, such as image classification \cite{perez2017effectiveness,mikolajczyk2018data,fawzi2016adaptive}, segmentation \cite{zhao2019data,ronneberger2015u}, object detection \cite{montserrat2017training,zhong2020random}, \etcno. Prevalent DA methods commonly edit samples in a random manner to generate virtual samples that do not actually exist in the collected training set. There is a series of works that augment samples with label-invariant transforms, \eg random rotation, flipping, erasing\cite{zhong2020random,devries2017improved}, \etcno. Other methods, such as Mixup~\cite{zhang2017mixup} and CutMix \cite{yun2019cutmix} combine different samples as well as their labels to synthesize virtual samples for augmentation. Besides, some automatic DA approaches \cite{cubuk2018autoaugment,lim2019fast,zhang2019adversarial,ho2019population,lin2019online,lin2021patch}  propose to search for more effective augmentation policy from a set of pre-designed DA operations.
\begin{figure}[t]
	\setlength{\abovecaptionskip}{0pt} 
\setlength{\belowcaptionskip}{-0pt}
	\begin{center}
		\includegraphics[width=1.0\linewidth]{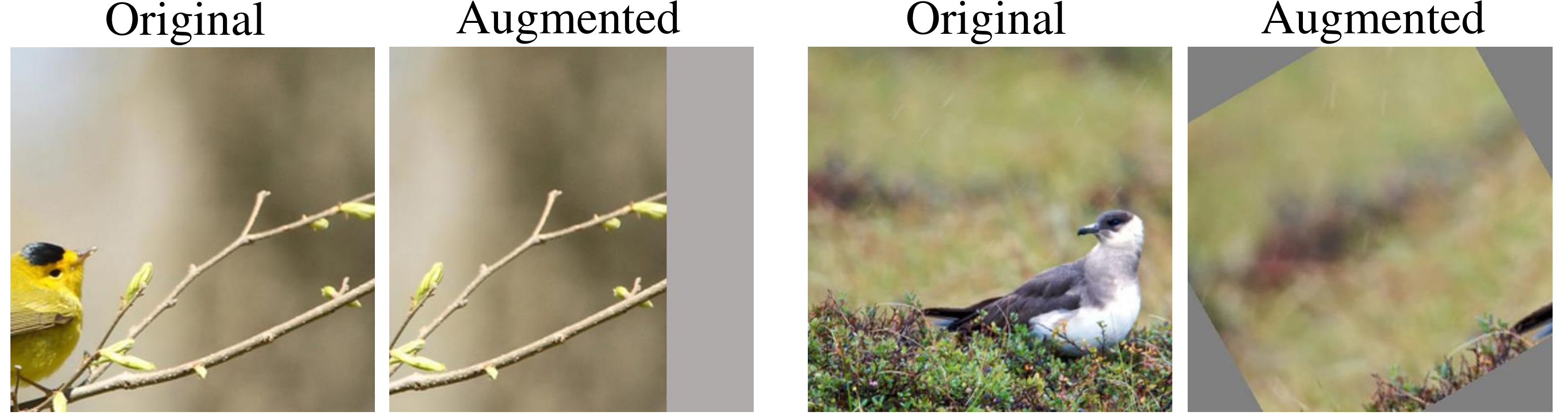}
	\end{center}
	\caption{Examples of the destructed samples when randomly selecting samples to augment (using AutoAugment \cite{cubuk2018autoaugment}).} 
	\label{fig:negative}
\vspace{-1.2em}
\end{figure}

\begin{figure}[t]
	\setlength{\abovecaptionskip}{0pt} 
\setlength{\belowcaptionskip}{-0pt}
	\begin{center}
		\includegraphics[width=1\linewidth]{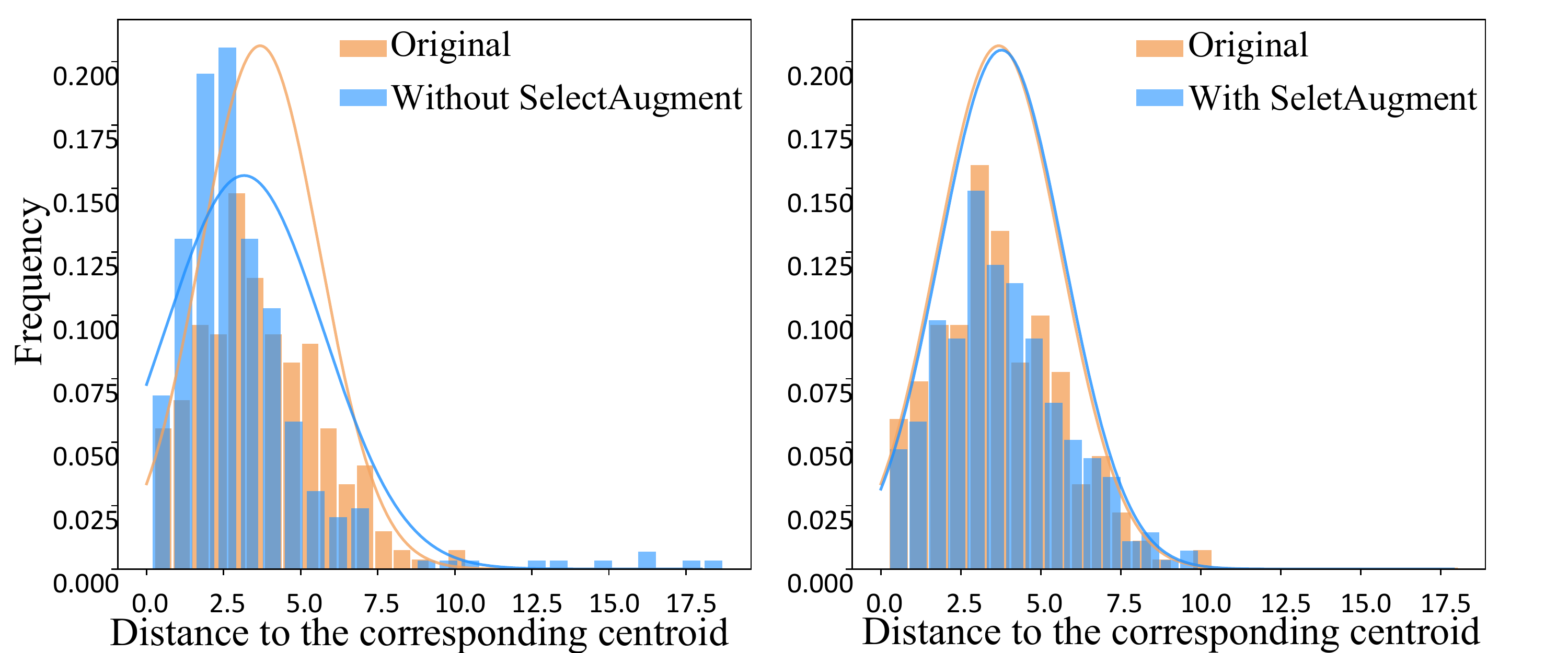}
	\end{center}
	\caption{Illustration of the existence of the distribution shift when using DA and the role of our proposed SelectAugment in alleviating this. Here, we compare the histograms of distances to the corresponding centroids between the augmented data without (left) and with (right) SelectAugment. }
	\label{fig:distribution}
\vspace{-1.5em}
\end{figure}

Prior works mentioned above only focus on how to augment but ignore the selection of samples suitable to be augmented. In other words, each training sample is randomly augmented with a probability that is usually heuristically set or automatically searched. However, randomly selecting samples for DA may cause content destruction and visual ambiguities as the examples shown in Fig.~\ref{fig:negative}, further leading to the distribution shift.
Here, we showcase this by comparing the feature distribution of original data to the augmented data with a representative DA method (\ieno, AutoAugment) in the left sub-figure of Fig.~\ref{fig:distribution}.
This uncontrollable shift on the training data may incur semantic ambiguities, and thus limit the benefits of DA, even lead to performance loss on the test data. A simple example is that adopting rotation augmentation on the number image ``6'' may cause confusion with ``9''.
Thus, it's of great importance to design a deterministic sample selection strategy for DA to eliminate such randomness, which is still under-explored.




One straightforward idea for addressing the above problem is to decide whether to augment an image or not in each batch in a deterministic way, which is in line with the decision-making problem in reinforcement learning (RL). 
However, directly applying RL will suffer from difficult optimization due to the large action space (\ieno, $2^{b}$ where $b$ denotes the size of mini-batch).
Towards easier policy learning, we propose a novel approach named SelectAugment to break the problem down into two steps and adopt Hierarchical Reinforcement Learning (HRL) to online learn a parent policy and a child policy for performing hierarchical deterministic sample selection for more effective DA.
To be clear, we name the model for DA strategy learning as ``policy network'' while the model for the mainstream task as ``target network''. The parent policy aims to first choose the augmentation ratio for each batch from a ratio pool $P$ according to sample information and the training status of the target network. Then, under this ratio, the child policy performs the sample-level selection for executing DA based on the contents and semantics of samples. 
Considering the hierarchy, the action space size of the parent policy is the size of ratio pool. For the action space of the child policy, its size is conditioned on the parent policy which is denoted as $C_b^{\lfloor{b{\times}{p}}\rfloor}$, where $p \in P$ and $C_n^m$ represents the number of combinations of $m$ selected from $n$. Easy to find that action spaces are effectively reduced compared to $2^b$. With a smaller action space, it's easier for us to find an effective strategy.

The parent policy and the child policy in our proposed approach are jointly optimized with the target network (\ieno, the mainstream task model) together, which leaves our proposed approach as an online one, obviating the need for re-training the target network after learning the sample selection strategy. In each training iteration, the selected images processed by off-the-shelf DA methods, such as Mixup \cite{zhang2017mixup}, CutMix \cite{yun2019cutmix} or AutoAugment \cite{cubuk2018autoaugment}, are combined together with the remaining images staying original to form a new batch that is fed into the target network for task training. We compare the calculated loss inferred by this newly constructed batch with the losses inferred by non-augmentation batch and fully-augmented batch respectively and take their difference as the reward signal to train the policy network. For the target network, we just use the batch augmented with our proposed SelectAugment for training. As illustrated in the right sub-figure of Fig.~\ref{fig:distribution}, our proposed method effectively prevents uncontrollable distribution shift when adopting data augmentation. Besides, extensive experiments demonstrate it's applicable for different off-the-shelf DA methods.


In summary, our contributions lie in three aspects: 
1) We are the first to pinpoint that deterministic sample selection matters in bringing different DA methods into their full play by reducing the side effects caused by the uncontrollable randomness of DA, which is overlooked in prior works;
2) We model the sample selection for DA as a two-step decision-making problem delicately and learn the policy via HRL, where we first determine the augmentation ratio at the batch level, then perform a deterministic allocation for executing augmentation operations at the instance level. 3) We conduct extensive experiments to demonstrate our proposed method can be generally applied to different off-the-shelf DA methods and enhance them consistently.

\section{Related Works}

\subsection{Data Augmentation}

Data augmentation (DA) plays a critical role in deep learning, which can effectively alleviate network overfitting problems.
Previous commonly used methods perform simple transformations, such as random rotation and translation \cite{simard2003best}.
CutOut \cite{devries2017improved} and its variants \cite{takahashi2019data,zhong2020random} randomly crop regions of an image to form stronger perturbations. 
Furthermore, recent progress in automated machine learning has began to study \cite{cubuk2018autoaugment,lim2019fast} automatically searching for the optimal transformation policy to relieve human expertise.
In addition, there are label-perturbing DA methods. For instance, Mixup \cite{zhang2017mixup} uses two training images in both pixel and label space. CutMix \cite{yun2019cutmix} randomly crops a region of one image and pastes it into another image, mixing labels with the proportion of two images.

These previous DA methods \cite{zhang2017mixup,yun2019cutmix,cubuk2018autoaugment} have achieved impressive results on many tasks, but they do not consider whether the samples are suitable for augmentation. In these methods, a probability is given for each image to roughly decide whether to augment or not, where the probability is commonly manually designed. For example, the value of probability is a fixed scalar in Mixup while a linearly increased scalar in CutMix. In particular, AutoAugment automatically searches for optimal augmentation policies including this aforementioned probability. To be specific, AutoAugment generates a tuple (operation, magnitude, probability) and evaluates it with the performance of a small proxy network that is trained from scratch. When retraining the target network, one of the offline policies is randomly selected and performed on the training data to train the target network. Therefore, they all ignore the image contents when using offline policies for DA, and decide whether to augment or not in a stochastic way. Recently, a series of online automatic DA methods \cite{zhang2019adversarial,lin2019online,lin2021patch} learns to choose the DA operations based on the sample contents, but still executes them with a manually set probability.

As long as the probability is used, no matter whether it is manually set or automatically searched, there is randomness for each specific image when determining whether to augment it or not.
As aforementioned, such randomness in selecting data for DA is of high risk to incur the optimization biases due to the introduced distribution shift. In contrast to these DA works that are stochastic, we propose a deterministic data selection approach to select the samples to execute the DA operations according to their contents and the network training status,  
which is generally applicable to different tasks to enlarge the benefits of DA.

\subsection{Hierarchical Reinforcement Learning}

Hierarchical RL (HRL) decomposes a complex task into several sub-ones with a hierarchical topology to speed up the learning process.
Specifically, goal-conditioned HRL \cite{levy2017learning,nachum2018data,kulkarni2016hierarchical} divides the corresponding task into two steps, which are respectively completed by two graded policies (\ieno, the parent and child policies) that are learned simultaneously.
Commonly, the parent policy outputs preliminary goals as the guidance for its subordinate child policy in a top-down view.
Some early works require manual design of goals \cite{peng2017composite,cuayahuitl2010evaluation}, while some methods\cite{florensa2018automatic,tang2018subgoal} (including our proposed method) automatically generate goals through the interaction with the environment. According to goals, second-step policy executes actions at a more granular level. Recent advanced goal-conditioned HRL works have been successfully applied to different tasks and achieved surprising results, such as open-domain dialog \cite{saleh2020hierarchical}, game playing \cite{kulkarni2016hierarchical}, interactive navigation \cite{li2020hrl4in}, course recommendation~ \cite{zhang2019hierarchical}.


\section{Proposed Method}
In this section, we first outline the framework of SelectAugment as well as its core idea, then elaborate on the hierarchical deterministic sample selection policy learning in SelectAugment. Furthermore, we discuss the superiority and an important expansion of our proposed method.


\vspace{-0.8em}
\begin{figure*}[h]
	\setlength{\abovecaptionskip}{0pt} 
\setlength{\belowcaptionskip}{-0pt}
	\begin{center}
		\includegraphics[width=1.0\linewidth]{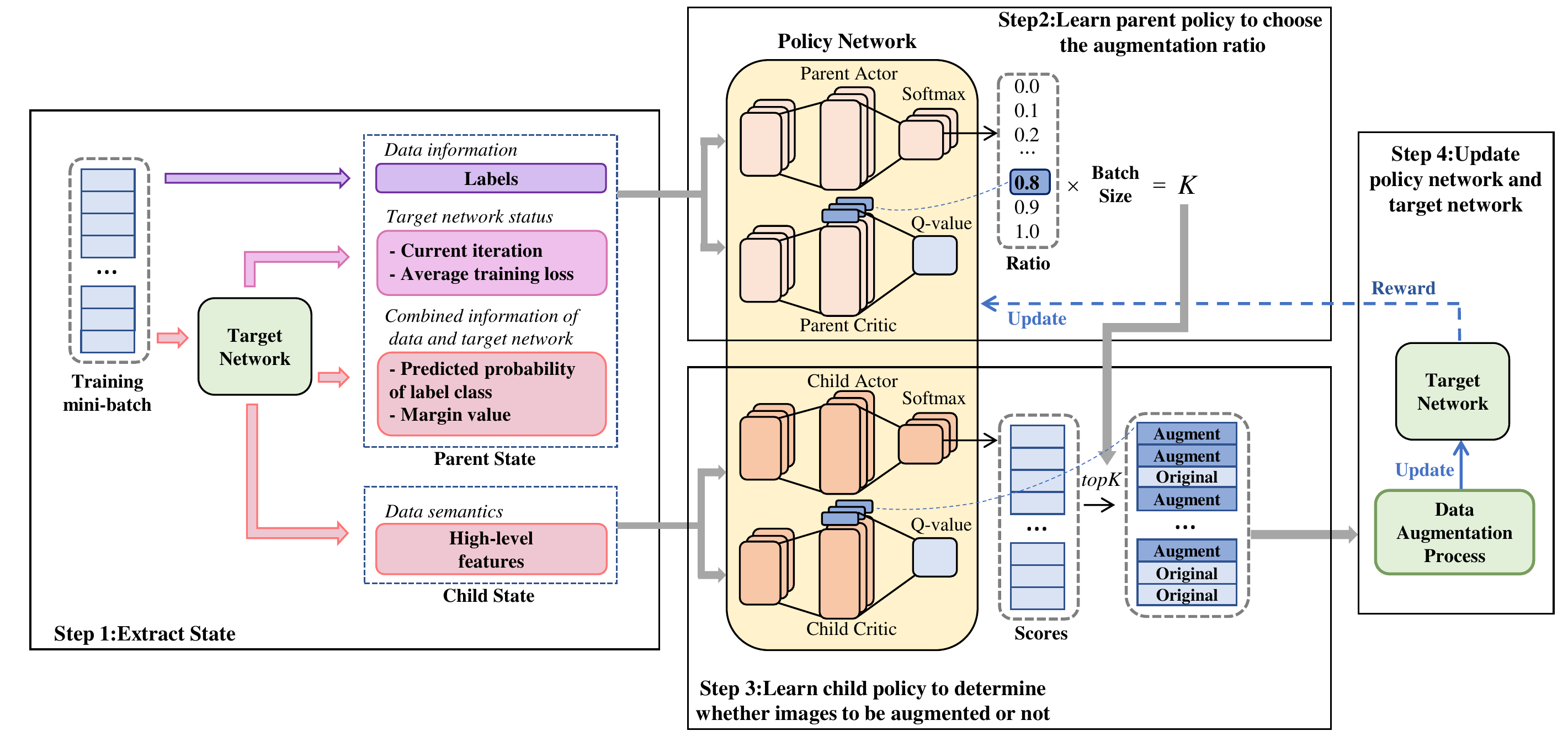}
	\end{center}
	\caption{The framework of our proposed method. Parent policy first selects the augmentation ratio of each batch. Then, child policy specifically determines which images to be augmented. }
	\label{fig:framework}
\vspace{-1.5em}
\end{figure*}

\subsection{Overview}
Applying data augmentation (DA) on different training samples is of diverse risks to incur semantic ambiguities, which brings our attention to the sample selection for DA. In this work, we propose SelectAugment, which aims to learn a sample selection policy for more effective DA. However, directly adopting RL for this objective encounters difficult optimization due to the large action space to be explored. Thereby, we decompose the problem into two-step and adopt Hierarchical RL to solve it. As shown in Fig. \ref{fig:framework}, an augmentation ratio is first determined by parent policy at the batch level according to the training status of the target network. Then, child policy sorts all samples in the current batch based on their contents and achieves sample selection under this ratio. Our proposed method can be used in tandem with various off-the-shelf DA designs. In the following, we shed light on the detailed formulation and modeling of our proposed SelectAugment.

\subsection{Basics of Reinforcement Learning}

We introduce the basics of Reinforcement Learning (RL) in this part. RL commonly model the policy learning problem as a Markov decision process (MDP) represented with $({\mathcal{S}},{\mathcal{A}},{\mathcal{P}},R,\gamma,T )$. Here, ${\mathcal{S}}$ and ${\mathcal{A}}$ denote the state space and the action space respectively. The RL agent observes the environment state $s \in {\mathcal{S}}$ and takes an action $a \in {\mathcal{A}}$ with the policy $\pi (a\left| {s)} \right.:{\mathcal{S}}\times {\mathcal{A}} \to [0,1]$. Then, the RL agent receives a step-wise reward $r:{\mathcal{S}} \times {\mathcal{A}} \to \mathbb{R}$. 
Accordingly, the environment moves to next state with a transition function denoted as ${\mathcal{P}}:{\mathcal{S}} \times {\mathcal{A}} \times {\mathcal{S}} \to [0,1]$.
$\gamma \in (0,1]$ is a discount factor and $T$ is a time horizon.
The objective of policy learning is to learn an optimal policy ${\pi ^*}$ which can maximize the accumulative reward $R$ over different steps in each episode.

Hierarchical Reinforcement Learning (HRL) aims to decompose a complex task into a hierarchy of several sub-tasks. Specifically, a two-level HRL commonly learns a parent policy ${\pi ^P}({a^P}\left| {{s^P}} \right.)$ and a child policy ${\pi ^C}({a^C}\left| {{s^C},{a^P}} \right.)$ corresponding to $MD{P^P}\!=\!({\mathcal{S}^P},{\mathcal{A}^P},{{\mathcal{P}^P}},{R^P},{\gamma},T)$ and $MD{P^C}\!=\!({{\mathcal{S}^C}},{{\mathcal{A}^C}},{{\mathcal{P}^C}},{R^C},{\gamma},T)$, respectively.
The ${{\mathcal{A}^P}}$ denotes the action space of parent policy while the ${{\mathcal{A}^C}}$ denotes the child action space. The parent policy outputs a parent action $a^P \in {{\mathcal{A}^P}}$, which is taken as the condition for the following decision by the child policy.

In this paper, we employ HRL to decide an augmentation ratio at the batch level and select specific samples to be augmented at the instance level.

\subsection{Hierarchical Deterministic Sample Selection Policy}

In our proposed SelectAugment, we formulate the task of data selection as a two-step decision-making problem and adopt Hierarchical Reinforcement Learning (HRL) to search for the optimal policy. We detail our design below.

\subsubsection{Parent Policy Modeling} 

The parent policy aims to determine the proportion of the augmented images in each batch. As illustrated in Fig. \ref{fig:framework}, we adopt RL to learn the policy towards this objective, and model the \textit{state} and \textit{action} for the policy learning of this level as below.

%
\textit{Parent state.}  
As illustrated in Fig. \ref{fig:framework}, we take the current batch and the status variables of the target network together as the parent state vector $\bm{s^P}$.
Specifically, $\bm{s^P}$ is required to represent not only the characteristics of the training batch, the training status of target network, but also the responses of the target network on the current batch.
To achieve this, the parent state vector $\bm{s^P}$ encodes three different categories of information in our design: 1) Encoding the information about the data itself, \ieno, labels (denoted by $\bm y$) of data, which represents the coarse semantics of the current batch. 2) Encoding the information about the target network, including the training iteration number and the recorded average historical training loss of target network. Here, we choose such variables to reflect the training status following \cite{fan2018learning,kumar2010self,jiang2014easy} where the effectiveness of such design has been experimentally demonstrated. 3) Encoding the feedback of the target network on the current batch, including the predicted probability of label class as well as a margin value proposed in ~\cite{cortes2013multi} with the definition of $P(\bm{y}\left| \bm{x} \right.) - {\max _{{\bm{y^{'}}} \ne \bm{y}}}P(\bm{{y^{'}}}\left| \bm{x} \right.)$.

\textit{Parent action.}
The parent policy determines what proportion of samples in the current batch to be augmented by choosing the augmentation ratio from the ratio pool. 
We set up the augmentation ratio pool (\ieno, the action space of parent agent) as $ {{\mathcal{A}^P}}={\rm{\{ 0.0,0}}{\rm{.1,0}}{\rm{.2,}}...{\rm{,0}}{\rm{.9,1.0\} }}$. Here, $a^P=0$ represents \textit{non-augmentation} in the sense that all images in the batch are original images, while $a^P=1$ denotes \textit{fully-augmentation} where all images in the batch are augmented. We discuss the impact of the configuration of augmentation ratio pool in the supplementary.




\subsubsection{Child Policy Modeling}
The child policy aims to determine whether each image in a batch is augmented or not under the learned ratio by parent policy, as shown in Fig. \ref{fig:framework}. This means that we make an instance-level decision on sample selection based on their contents with the child policy.

%
\textit{Child state.}
The child policy needs to perceive the sample contents for finding the most suitable samples to be augmented under the augmentation ratio inferred by parent policy. 
Therefore, we propose to utilize the deep features of images extracted by target network as child state $\bm{s^C}$.


\textit{Child action.}
Different from the parent policy that determines the augmentation ratio in a coarse way, the child policy aims to make a precise decision on whether each image is augmented. Here, we define child action $\mathbf{a} ^C$ as a vector whose dimension equals to the batch-size, which represents the scores of images suitable for augmentation in each batch. Given an augmentation ratio $a^P$ inferred by the parent policy, the number of augmented images $K$ can be determined as $ K=\lfloor a^P \times b \rfloor$ where $b$ is the size of mini-batch.
Then, child policy selects the samples corresponding to the $K$-highest scores $topK(\cdot)$ to execute augmentation operations $\psi(\cdot)$ as Eq.( \ref{equ:topK}).
\begin{equation}
  \bm{x_{aug}}\!=\!\{ \psi(x_i)|a_i^C \in topK({\mathbf{a^C}}),x{_i} \in \bm{x})\},
  \label{equ:topK}
\end{equation}
\begin{equation}
  \bm{x_{ori}} =\!\{ x_j|a_j^C \notin topK({\mathbf{a^C}}),x{_j} \in \bm{x}\},
  \label{equ:ori}
\end{equation}
where $\bm{x}$ denotes the original training mini-batch (without any augmentation operations) and $\bm{\widetilde x}=\{ \bm{x_{aug}},\bm{x_{ori}}\}$ represents the selectively augmented mini-batch processed by SelectAugment.






\subsubsection{Reward Function}
The objective of our proposed data selection policy is to enhance the mainstream task through improving the effectiveness/benefits of DA as possible. 
Therefore, we compare the feedback of target network $\phi(\cdot)$ on the selectively augmented data processed by our proposed SelectAugment $\bm{\widetilde x}$ with on 1) the original data $\bm{x}$ and 2) the fully-augmented data $\bm{\overline x}$ respectively, and take their differences on the training losses as the reward for the policy network learning in RL. Mathematically, the reward function can be formulated as:
\begin{equation}
{r} =  l(\phi (\bm{x}),\bm{y})-l(\phi (\bm{\widetilde x}),\bm{y}) + l(\phi (\bm{\overline x} ),\bm{y})-l(\phi (\bm{\widetilde x}),\bm{y}),  
\label{equ:reward}
\end{equation}

The parent policy and the child policy serve for the same goal, thus, we adopt the same reward function design as the Eq.\ref{equ:reward} for these two levels of RL tasks.


\subsubsection{Hierarchical Policy Learning}
In this part, we introduce the hierarchical training for the parent policy and the child policy stated above.
Considering the action spaces of both the parent policy and the child policy are discrete, in this work, we adopt the widely used Advantage Actor-Critic (A2C) algorithm \cite{mnih2016asynchronous} to perform the hierarchical policy learning. Specifically, there are two neural networks in the Actor-Critic algorithm, where the actor network is to learn a discrete control policy ${\pi}(a\left| s \right.)$ while the critic network aims to estimate the value of state $V^\pi(s)$. 
Here, we reformulate it appropriately under our task scenario. Similar to \cite{mnih2016asynchronous}, we model the value of each state-action pair with a Q-value, which is formulated as: 
\begin{equation}
{Q^\pi}(a^P,\bm{a^C},\bm{s^C},\bm{s^P})\!=\!{\mathbb{E}_\pi }[ {{r}\left| {{\mathcal{A}}\!=\!(a^P,\mathbf{a^C}),{\mathcal{S}}\!=\!(\bm{s^P}, \bm{s^C)}} \right.} ].
\end{equation}
The model for the parent policy learning comprises a parent actor network and a parent critic network as shown in Fig.\ref{fig:framework}. We use $\theta_{P}$ and $\varphi_{P}$ to denote their corresponding trainable network parameters, respectively. Here, we define the  advantage function of parent policy for updating $\theta_{P}$ and $\varphi_{P}$:
\begin{equation}
{A_P}(a^P,\bm{s^P}) = {Q^\pi }(a^P,\mathbf{a^C},\bm{s^C},\bm{s^P}) - V_{\varphi_{P}}^{\pi_{\theta_{P}}}(\bm{s^P}).
\end{equation}
Specifically, we take the square value of the advantage function as the loss function to update the $\varphi_{P}$:

\begin{equation}
L(\varphi_{P}) = {\left( A_P(a^P,\bm{s^P}) \right)^2}.
\end{equation}
Moreover, the loss function for updating the $\theta_{P}$ is:
\begin{equation}
{L}(\theta_{P}) =  - \log {\pi _{{\theta_{P}}}}(a^P|\bm{s^P})A_P(a^P,\bm{s^P}).
\end{equation}
Similarly, the model for the child policy learning comprises a child actor (with parameters $\theta_{C}$) and a child critic (with parameters $\varphi_{C}$). The advantage function of the child policy and the loss function for updating the $\varphi_{C}$ are defined as:
\begin{equation}
{A_C}(\mathbf{a^C},\bm{s^P}) = {Q^\pi }(a^P,\mathbf{a^C},\bm{s^C},\bm{s^P}) - V_{\varphi_{C}}^{\pi_{\theta_{C}}}(\bm{s^C}),
\end{equation}
\begin{equation}
L({\varphi_{C}}) = {\left( A_C(\mathbf{a^C},\bm{s^P}) \right)^2}.
\end{equation}
Inspired by top-k policy \cite{chen2019top}, the loss function used to update $\theta_{C}$ is modified to the following:
\begin{equation}
L(\theta_{C}) =  - \sum\nolimits_{a^{C}_{i} \in {topK{(\bm{a^C)}}}} {\log {\pi _{{\theta _{C}}}}(a^{C}_{i}|{\bm{s^C}})A_C({\mathbf{a^C}},{\bm{s^C}})}.
\end{equation}



\subsection{Discussion}

\subsubsection{The superiority on the action space}


The action space of RL algorithm refers to the set including all possible actions.
For the data selection problem we tackle in this paper, a straightforward solution is to independently determine whether to execute the DA operation or not for each sample in a batch, which makes the size of the action space be $2^{b}$ ($b$ is the batch size). A large action commonly renders difficult policy exploration.
Thanks to the hierarchy of our proposed SelectAugment based on HRL, the action space size is effectively reduced.
In our design, the action space size of the parent policy is the size of the ratio pool $|{\mathcal{A}}^P|$. For the action space of child policy, its size is conditioned on the decision of the parent policy, denoted as $C_b^{\lfloor{b{\times}{a^P}}\rfloor}$ where $C_n^m$ represents the number of combinations of $m$ selected from $n$. Note that $C_b^{\lfloor{b{\times}{a^P}}\rfloor}\le C_b^{\frac{b}{2}}<2{^b}$. To be more intuitive, we take the batch-size is $256$ as an example. Even in the extreme case (\ieno, the ratio given by the parent policy is $0.5$), the action space of child policy is up to the largest, which still achieves $20$ times reduction compared to $2^b$.

Therefore, the size of the action space of our proposed SelectAugment is effectively reduced by decomposing the data selection problem into a two-step decision processing. 
As explained in \cite{zahavy2018learn,wei2018hierarchical}, a smaller action space is beneficial to reduce the difficulty of policy exploration. Furthermore, we experimentally demonstrate this improves the performance of the target network in the following.
\subsubsection{An extension \textit{SelectAugment+}}
At the heart of our work is to select suitable samples for executing off-the-shelf DA operations. Note that it has the general applicability to different DA operations proposed in prior works, such as Mixup \cite{zhang2017mixup}, Cutmix \cite{yun2019cutmix}, AutoAugment \cite{cubuk2018autoaugment}, Cutout \cite{devries2017improved}, \etcno. Beyond this, in the supplementary, we introduce an extension ``SelectAugment+'' in which we also take the content-adaptive DA operation selection into our consideration. In other words, we further learn how to select the most suitable DA operations for the samples selected by our proposed SelectAugement. Please refer to the supplementary for the detailed introduction and the experimental results.

\section{Experiments and Results}

\subsection{Experimental Settings}

\subsubsection{Datasets} To our best knowledge, image classification is the most widely used task for validating the effectiveness of DA methods \cite{shorten2019survey,zhang2017mixup,yun2019cutmix,cubuk2018autoaugment}. 
We thus evaluate our SelectAugment on CIFAR-10~\cite{krizhevsky2009learning}, CIFAR-100 ~\cite{krizhevsky2009learning} and ImageNet~\cite{deng2009imagenet}.
Besides, we further evaluate our proposed method on fine-grained image recognition. Here we use CUB-200-2011~\cite{wah2011caltech} and Stanford Dogs~\cite{khosla2011novel} for evaluation. The detailed dataset introduction is placed in the supplementary.

\vspace{-0.1em}
\subsubsection{Implementation Details} 
Following \cite{cubuk2018autoaugment,lim2019fast,lin2021patch,yun2019cutmix}, for CIFAR-10 and CIFAR-100, we respectively use Wide-ResNet-28-10 (WRN-28-10) \cite{zagoruyko2016wide}, ShakeShake(26 2$\times$32d) and ShakeShake (26 2$\times$96d) \cite{gastaldi2017shake} (aliased as SS(32$\times$d) and SS(96$\times$d)) as target network.
For ImageNet \cite{deng2009imagenet},
ResNet-50 and ResNet-200~\cite{he2016deep} are adopt as target models, which are trained from scratch. For fine-grained image classification,  we follow the previous works  \cite{du2020fine,chen2019destruction} and utilize pre-trained ResNet-50 and ResNet-152 models as the target network.  

In all experiments, the parent policy network is consist of fully convolutional networks (FCNs) to output the value of augmentation ratio, and the child policy network is a multi-layer convolutional neural network
(CNN), which is employed to output scores of images. We set batch size to $128$ for experiments on CIFAR and $256$ for experiments on ImageNet together with two fine-grained datasets. Other hyperparameters, such as training epochs and the learning rate of target network are the same as previous works \cite{yun2019cutmix,zhang2017mixup,cubuk2018autoaugment} for fair comparison and are reported in the supplementary material. To ensure the reliability of our experiments, we conduct the same experiments four times with different random seeds. 
\begin{table}[t]
\vspace{-0.5em}
\setlength{\abovecaptionskip}{0.3cm}
\begin{center}
\captionsetup{width=1.0\linewidth}
\caption{\textbf{CIFAR-10 and CIFAR-100 results.} We report test accuracy ($\%$) (average $\pm$ standard deviations (std) of four runs) of various augmentation techniques using different data selection policies. AutoAug is short for AutoAugment.}
\label{table:CIFAR}
\scalebox{0.71}{
\begin{tabular}{c|c|cc|cc}
\hline \hline
\multicolumn{1}{c|}{\multirow{2}{*}{DA}}                                                                & \multicolumn{1}{c|}{\multirow{2}{*}{Policy}} & \multicolumn{2}{c|}{CIFAR-10}   & \multicolumn{2}{c}{CIFAR-100}   \\ \cline{3-6} 
\multicolumn{1}{c|}{}                                                                                   & \multicolumn{1}{c|}{}                        & WRN-28-10     &  \multicolumn{1}{c|} {SS(32$\times$d)}      & WRN-28-10      & SS(96$\times$d)      \\ \hline
\multicolumn{2}{c|}{Baseline}                                                                                                                           & 96.13\scriptsize$\bm\pm 0.13$         & 96.26\scriptsize$\bm\pm 0.17$        & 81.20\scriptsize$\bm\pm 0.10$         & 82.85\scriptsize$\bm\pm 0.11$          \\ \hline
\multicolumn{1}{c|}{\multirow{5}{*}[-0.1cm]{Mixup}}                                                            & all                                          & 97.14\scriptsize$\bm\pm 0.11$          & 97.12\scriptsize$\bm\pm 0.09$          & 82.41\scriptsize$\bm\pm 0.13$          & 84.77\scriptsize$\bm\pm 0.15$          \\
\multicolumn{1}{c|}{}                                                                                   & random                                       & 96.85\scriptsize$\bm\pm 0.22$          & 96.91\scriptsize$\bm\pm 0.13$          & 82.25\scriptsize$\bm\pm 0.21$          & 84.68\scriptsize$\bm\pm 0.17$          \\
\multicolumn{1}{c|}{}                                                                                   & fixed                                           & 96.88\scriptsize$\bm\pm 0.09$          & 96.99\scriptsize$\bm\pm 0.11$          & 82.87\scriptsize$\bm\pm 0.16$          & 84.79\scriptsize$\bm\pm 0.14$          \\
\multicolumn{1}{c|}{}                                                                                   & linearly                                     & 96.89\scriptsize$\bm\pm 0.07$          & 96.95\scriptsize$\bm\pm 0.10$          & 82.60\scriptsize$\bm\pm 0.18$          & 84.56\scriptsize$\bm\pm 0.13$          \\

\multicolumn{1}{c|}{}                                                                                   & onlineRL                                           & 96.99\scriptsize$\bm\pm 0.08$          & 97.05\scriptsize$\bm\pm 0.09$          & 82.91\scriptsize$\bm\pm 0.09$          & 85.09\scriptsize$\bm\pm 0.12$          \\
\multicolumn{1}{c|}{}                                                                                   & \textbf{Ours}                                & \textbf{97.33}\scriptsize$\mathbf{\bm\pm 0.06}$ & \textbf{97.38}\scriptsize$\mathbf{\bm\pm 0.07}$ & \textbf{83.37}\scriptsize$\mathbf{\bm\pm 0.11}$ & \textbf{85.17}\scriptsize$\mathbf{\bm\pm 0.10}$ \\ \hline
\multicolumn{1}{c|}{\multirow{5}{*}[-0.1cm]{Cutmix}}                                                           & all                                          & 97.11\scriptsize$\bm\pm 0.05$          & 97.07\scriptsize$\bm\pm 0.07$          & 82.91\scriptsize$\bm\pm 0.14$          & 84.85\scriptsize$\bm\pm 0.10$          \\
\multicolumn{1}{c|}{}                                                                                   & random                                       & 97.05\scriptsize$\bm\pm 0.18$          & 96.95\scriptsize$\bm\pm 0.21$          & 82.83\scriptsize$\bm\pm 0.17$          & 84.89\scriptsize$\bm\pm 0.16$          \\
\multicolumn{1}{c|}{}                                                                                   & fixed                                           & 96.92\scriptsize$\bm\pm 0.11$          & 96.99\scriptsize$\bm\pm 0.14$          & 82.87\scriptsize$\bm\pm 0.12$          & 84.92\scriptsize$\bm\pm 0.13$          \\
\multicolumn{1}{c|}{}                                                                                   & linearly                                     & 97.24\scriptsize$\bm\pm 0.07$          & 97.21\scriptsize$\bm\pm 0.05$          & 83.12\scriptsize$\bm\pm 0.16$          & 84.97\scriptsize$\bm\pm 0.18$          \\
\multicolumn{1}{c|}{}                                                                                   & onlineRL                                           & 96.99\scriptsize$\bm\pm 0.06$          & 97.03\scriptsize$\bm\pm 0.10$          & 82.95\scriptsize$\bm\pm 0.14$          & 85.01\scriptsize$\bm\pm 0.12$          \\
\multicolumn{1}{c|}{}                                                                                   & \textbf{Ours}                                & \textbf{97.41}\scriptsize$\mathbf{\bm\pm 0.07}$ & \textbf{97.44}\scriptsize$\mathbf{\bm\pm 0.06}$ & \textbf{83.45}\scriptsize$\mathbf{\bm\pm 0.14}$ & \textbf{85.24}\scriptsize$\mathbf{\bm\pm 0.07}$  \\ \hline
\multicolumn{1}{c|}{\multirow{5}{*}[-0.1cm]{AutoAug}} & all                                          & 97.28\scriptsize$\bm\pm 0.08$          & 97.37\scriptsize$\bm\pm 0.14$          & 83.08\scriptsize$\bm\pm 0.17$          & 85.66\scriptsize$\bm\pm 0.11$          \\
\multicolumn{1}{c|}{}                                                                                   & random                                       & 97.13\scriptsize$\bm\pm 0.29$          & 97.23\scriptsize$\bm\pm 0.22$          & 83.23\scriptsize$\bm\pm 0.34$          & 84.75\scriptsize$\bm\pm 0.27$          \\
\multicolumn{1}{c|}{}                                                                                   & fixed                                           & 97.19\scriptsize$\bm\pm 0.17$          & 97.14\scriptsize$\bm\pm 0.21$          & 83.20\scriptsize$\bm\pm 0.16$          & 85.37\scriptsize$\bm\pm 0.18$          \\
\multicolumn{1}{c|}{}                                                                                   & linearly                                     & 96.92\scriptsize$\bm\pm 0.11$          & 96.91\scriptsize$\bm\pm 0.21$          & 83.17\scriptsize$\bm\pm 0.25$          & 85.52\scriptsize$\bm\pm 0.18$          \\
\multicolumn{1}{c|}{}                                                                                   & onlineRL                                           & 97.34\scriptsize$\bm\pm 0.22$          & 97.37\scriptsize$\bm\pm 0.12$          & 83.14\scriptsize$\bm\pm 0.17$          & 85.61\scriptsize$\bm\pm 0.16$          \\
\multicolumn{1}{c|}{}                                                                                   & \textbf{Ours}                                & \textbf{97.65}\scriptsize$\mathbf{\bm\pm 0.12}$ & \textbf{97.61}\scriptsize$\mathbf{\bm\pm 0.09}$ & \textbf{83.81}\scriptsize$\mathbf{\bm\pm 0.12}$ & \textbf{85.87}\scriptsize$\mathbf{\bm\pm 0.13}$ \\ \hline
\hline
\end{tabular}}
\end{center}
\vspace{-1.8em}
\end{table}

\begin{table}[t]
\vspace{0.4em}
\setlength{\abovecaptionskip}{0.3cm}
\begin{center}
\captionsetup{width=1.0\linewidth}
\caption{\textbf{ImageNet results.} We report Top-1 / Top-5 validation accuracy ($\%$). Performance is the average $\pm$ std.}
\label{table:imagenet}
\scalebox{0.71}{
\begin{tabular}{cc|c|c}
\hline \hline
\multicolumn{1}{c|}{\multirow{2}{*}{DA}}           & \multicolumn{1}{c|}{\multirow{2}{*}{Policy}} & \multicolumn{2}{c}{ImageNet}                   \\ \cline{3-4} 
\multicolumn{1}{c|}{}                              & \multicolumn{1}{c|}{}                        & ResNet-50               & ResNet-200                 \\ \hline
\multicolumn{2}{c|}{Baseline}                                                                      & 76.28{\scriptsize$\bm\pm 0.15$} / 93.05{\scriptsize$\bm\pm 0.12$}          & 78.47{\scriptsize$\bm\pm 0.11$} / 94.19{\scriptsize$\bm\pm 0.17$}             \\ \hline
\multicolumn{1}{c|}{\multirow{5}{*}[-0.1cm]{Mixup}}       & all                                          & 77.01{\scriptsize$\bm\pm 0.20$} / 93.43{\scriptsize$\bm\pm 0.16$}          & 79.62{\scriptsize$\bm\pm 0.22$} / 94.83{\scriptsize$\bm\pm 0.15$}                \\
\multicolumn{1}{c|}{}                              & random                                       & 76.86{\scriptsize$\bm\pm 0.22$} / 93.21{\scriptsize$\bm\pm 0.19$}          & 79.48{\scriptsize$\bm\pm 0.25$} / 94.71{\scriptsize$\bm\pm 0.20$}             \\
\multicolumn{1}{c|}{}                              & fixed                                           & 76.82{\scriptsize$\bm\pm 0.10$} / 93.24{\scriptsize$\bm\pm 0.13$}          & 79.46{\scriptsize$\bm\pm 0.16$} / 93.66{\scriptsize$\bm\pm 0.14$}            \\
\multicolumn{1}{c|}{}                              & linearly                                     & 76.95{\scriptsize$\bm\pm 0.12$} / 93.33{\scriptsize$\bm\pm 0.11$}          & 79.53{\scriptsize$\bm\pm 0.15$} / 94.72{\scriptsize$\bm\pm 0.13$}              \\
\multicolumn{1}{c|}{}                              & onlineRL                                           & 77.17{\scriptsize$\bm\pm 0.14$} / 93.46{\scriptsize$\bm\pm 0.10$}          & 79.64{\scriptsize$\bm\pm 0.16$} / 94.80{\scriptsize$\bm\pm 0.15$}             \\
\multicolumn{1}{c|}{}                              & \textbf{Ours}                                & \textbf{77.84{\scriptsize$\mathbf{\bm\pm 0.17}$} / 93.78{\scriptsize$\mathbf{\bm\pm 0.12}$}} & \textbf{80.42{\scriptsize$\mathbf{\bm\pm 0.15}$} / 95.26{\scriptsize$\mathbf{\bm\pm 0.11}$}}   \\ \hline
\multicolumn{1}{c|}{\multirow{5}{*}[-0.1cm]{Cutmix}}      & all                                          & 77.18{\scriptsize$\bm\pm 0.16$} / 93.44{\scriptsize$\bm\pm 0.11$}          & 79.81{\scriptsize$\bm\pm 0.18$} / 94.86{\scriptsize$\bm\pm 0.10$}            \\
\multicolumn{1}{c|}{}                              & random                                       & 76.76{\scriptsize$\bm\pm 0.23$} / 93.02{\scriptsize$\bm\pm 0.19$}          & 79.64{\scriptsize$\bm\pm 0.26$} / 94.78{\scriptsize$\bm\pm 0.22$}              \\
\multicolumn{1}{c|}{}                              & fixed                                           & 76.83{\scriptsize$\bm\pm 0.14$} / 93.26{\scriptsize$\bm\pm 0.06$}          & 79.84{\scriptsize$\bm\pm 0.15$} / 94.82{\scriptsize$\bm\pm 0.13$}              \\
\multicolumn{1}{c|}{}                              & linearly                                     & 77.23{\scriptsize$\bm\pm 0.12$} / 93.54{\scriptsize$\bm\pm 0.08$}          & 79.92{\scriptsize$\bm\pm 0.12$} / 94.90{\scriptsize$\bm\pm 0.14$}              \\
\multicolumn{1}{c|}{}                              & onlineRL                                           & 77.15{\scriptsize$\bm\pm 0.16$} / 93.39{\scriptsize$\bm\pm 0.11$}          & 79.94{\scriptsize$\bm\pm 0.15$} / 94.86{\scriptsize$\bm\pm 0.12$}             \\
\multicolumn{1}{c|}{}                              & \textbf{Ours}                                & \textbf{78.02{\scriptsize$\mathbf{\bm\pm 0.16}$} / 93.90{\scriptsize$\mathbf{\bm\pm 0.13}$}} & \textbf{80.66{\scriptsize$\mathbf{\bm\pm 0.18}$} / 95.31{\scriptsize$\mathbf{\bm\pm 0.12}$}}  \\ \hline
\multicolumn{1}{c|}{\multirow{5}{*}[-0.1cm]{AutoAug}} & all                                          & 77.61{\scriptsize$\bm\pm 0.17$} / 93.82{\scriptsize$\bm\pm 0.12$}          & 79.96{\scriptsize$\bm\pm 0.14$} / 95.02{\scriptsize$\bm\pm 0.15$}            \\
\multicolumn{1}{c|}{}                              & random                                       & 76.75{\scriptsize$\bm\pm 0.23$} / 92.83{\scriptsize$\bm\pm 0.18$}          & 79.82{\scriptsize$\bm\pm 0.22$} / 94.85{\scriptsize$\bm\pm 0.25$}           \\
\multicolumn{1}{c|}{}                              & fixed                                           & 77.63{\scriptsize$\bm\pm 0.14$} / 93.77{\scriptsize$\bm\pm 0.11$}          & 79.92{\scriptsize$\bm\pm 0.12$} / 94.96{\scriptsize$\bm\pm 0.13$}             \\
\multicolumn{1}{c|}{}                              & linearly                                     & 77.13{\scriptsize$\bm\pm 0.12$} / 93.14{\scriptsize$\bm\pm 0.11$}          & 80.03{\scriptsize$\bm\pm 0.18$} / 95.04{\scriptsize$\bm\pm 0.15$}               \\
\multicolumn{1}{c|}{}                              & onlineRL                                           & 77.32{\scriptsize$\bm\pm 0.18$} / 93.56{\scriptsize$\bm\pm 0.14$}          & 80.15{\scriptsize$\bm\pm 0.19$} / 95.08{\scriptsize$\bm\pm 0.17$}             \\
\multicolumn{1}{c|}{}                              & \textbf{Ours}                                & \textbf{78.16{\scriptsize$\mathbf{\bm\pm 0.16}$} / 93.97{\scriptsize$\mathbf{\bm\pm 0.11}$}} & \textbf{80.78{\scriptsize$\mathbf{\bm\pm 0.14}$} / 95.36{\scriptsize$\mathbf{\bm\pm 0.12}$}}  \\ \hline \hline
\end{tabular}}
\end{center}
\vspace{-1.2em}
\end{table}

\begin{table}[t]
\vspace{-0.5em}
\setlength{\abovecaptionskip}{0.3cm}
\begin{center}
\captionsetup{width=1.0\linewidth}
\caption{\textbf{Fine-grained classification results.} We report test accuracy ($\%$) (average($\pm$ std) across four runs.}
\label{table:fine-grained}
\scalebox{0.71}{
\begin{tabular}{cc|cc|cc}
\hline \hline
\multicolumn{1}{c|}{\multirow{2}{*}{DA}}           & \multicolumn{1}{c|}{\multirow{2}{*}{Policy}}                   & \multicolumn{2}{c|}{CUB-200-2011} & \multicolumn{2}{c}{Stanford Dogs} \\ \cline{3-6} 
\multicolumn{1}{c|}{}                              & \multicolumn{1}{c|}{}                            & ResNet-50       & ResNet-152      & ResNet-50       & ResNet-152      \\ \hline
\multicolumn{2}{c|}{Baseline}                                                                          & 75.23{\scriptsize$\bm\pm 0.15$}           & 79.75{\scriptsize$\bm\pm 0.12$}           & 85.04{\scriptsize$\bm\pm 0.18$}           & 88.49{\scriptsize$\bm\pm 0.14$}           \\ \hline
\multicolumn{1}{c|}{\multirow{5}{*}[-0.1cm]{Mixup}}       & all                                              & 78.79{\scriptsize$\bm\pm 0.14$}           & 81.41{\scriptsize$\bm\pm 0.13$}           & 87.47{\scriptsize$\bm\pm 0.17$}           & 89.60{\scriptsize$\bm\pm 0.16$}           \\
\multicolumn{1}{c|}{}                              & random                                          & 77.96{\scriptsize$\bm\pm 0.25$}           & 81.26{\scriptsize$\bm\pm 0.18$}           & 86.83{\scriptsize$\bm\pm 0.22$}           & 88.74{\scriptsize$\bm\pm 0.15$}           \\
\multicolumn{1}{c|}{}                              & fixed                                                & 78.77{\scriptsize$\bm\pm 0.11$}           & 81.58{\scriptsize$\bm\pm 0.12$}           & 87.31{\scriptsize$\bm\pm 0.13$}           & 88.79{\scriptsize$\bm\pm 0.12$}           \\
\multicolumn{1}{c|}{}                              & linearly                                      & 78.88{\scriptsize$\bm\pm 0.12$}           & 81.52{\scriptsize$\bm\pm 0.10$}           & 87.22{\scriptsize$\bm\pm 0.13$}           & 88.93{\scriptsize$\bm\pm 0.14$}           \\
\multicolumn{1}{c|}{}                              & onlineRL                                               & 78.91{\scriptsize$\bm\pm 0.15$}           & 81.70{\scriptsize$\bm\pm 0.15$}           & 87.45{\scriptsize$\bm\pm 0.17$}           & 88.99{\scriptsize$\bm\pm 0.13$}           \\
\multicolumn{1}{c|}{}                              & \textbf{Ours}                            & \textbf{80.94}\scriptsize$\mathbf{\bm\pm 0.16}$  & \textbf{82.75}\scriptsize$\mathbf{\bm\pm 0.14}$  & \textbf{88.12}\scriptsize$\mathbf{\bm\pm 0.18}$  & \textbf{90.40}\scriptsize$\mathbf{\bm\pm 0.16}$  \\ \hline
\multicolumn{1}{c|}{\multirow{5}{*}[-0.1cm]{Cutmix}}      & all                                              & 76.68{\scriptsize$\bm\pm 0.17$}           & 79.86{\scriptsize$\bm\pm 0.14$}           & 87.27{\scriptsize$\bm\pm 0.16$}           & 89.55{\scriptsize$\bm\pm 0.14$}           \\
\multicolumn{1}{c|}{}                              & random                                        & 76.62{\scriptsize$\bm\pm 0.23$}           & 78.90{\scriptsize$\bm\pm 0.14$}           & 87.43{\scriptsize$\bm\pm 0.22$}           & 89.43{\scriptsize$\bm\pm 0.16$}           \\
\multicolumn{1}{c|}{}                              & fixed                                            & 76.76{\scriptsize$\bm\pm 0.10$}           & 79.35{\scriptsize$\bm\pm 0.11$}           & 87.48{\scriptsize$\bm\pm 0.12$}           & 89.50{\scriptsize$\bm\pm 0.11$}           \\
\multicolumn{1}{c|}{}                              & linearly                                      & 76.91{\scriptsize$\bm\pm 0.12$}           & 79.41{\scriptsize$\bm\pm 0.13$}           & 87.61{\scriptsize$\bm\pm 0.14$}           & 89.64{\scriptsize$\bm\pm 0.11$}           \\
\multicolumn{1}{c|}{}                              & onlineRL                                            & 76.98{\scriptsize$\bm\pm 0.16$}           & 79.77{\scriptsize$\bm\pm 0.14$}           & 87.60{\scriptsize$\bm\pm 0.15$}           & 89.58{\scriptsize$\bm\pm 0.12$}           \\
\multicolumn{1}{c|}{}                              & \textbf{Ours}                               & \textbf{77.53}\scriptsize$\mathbf{\bm\pm 0.15}$  & \textbf{80.21}\scriptsize$\mathbf{\bm\pm 0.16}$  & \textbf{88.42}\scriptsize$\mathbf{\bm\pm 0.17}$  & \textbf{90.30}\scriptsize$\mathbf{\bm\pm 0.14}$  \\ \hline
\multicolumn{1}{c|}{\multirow{5}{*}[-0.1cm]{AutoAug}} & all                                           & 78.34{\scriptsize$\bm\pm 0.14$}           & 81.17{\scriptsize$\bm\pm 0.12$}           & 86.90{\scriptsize$\bm\pm 0.14$}           & 89.63{\scriptsize$\bm\pm 0.11$}           \\
\multicolumn{1}{c|}{}                              & random                                           & 78.31{\scriptsize$\bm\pm 0.23$}           & 81.25{\scriptsize$\bm\pm 0.18$}           & 86.93{\scriptsize$\bm\pm 0.24$}           & 89.69{\scriptsize$\bm\pm 0.25$}           \\
\multicolumn{1}{c|}{}                              & fixed                                              & 79.18{\scriptsize$\bm\pm 0.11$}           & 81.04{\scriptsize$\bm\pm 0.12$}           & 87.73{\scriptsize$\bm\pm 0.10$}           & 89.74{\scriptsize$\bm\pm 0.08$}           \\
\multicolumn{1}{c|}{}                              & linearly                                       & 79.06{\scriptsize$\bm\pm 0.13$}           & 80.96{\scriptsize$\bm\pm 0.12$}           & 86.87{\scriptsize$\bm\pm 0.14$}           & 89.82{\scriptsize$\bm\pm 0.14$}           \\
\multicolumn{1}{c|}{}                              & onlineRL                                              & 79.35{\scriptsize$\bm\pm 0.20$}           & 81.39{\scriptsize$\bm\pm 0.18$}           & 87.04{\scriptsize$\bm\pm 0.22$}           & 89.84{\scriptsize$\bm\pm 0.15$}           \\
\multicolumn{1}{c|}{}                              & \textbf{Ours}                               & \textbf{81.27}\scriptsize$\mathbf{\bm\pm 0.16}$  & \textbf{82.45}\scriptsize$\mathbf{\bm\pm 0.15}$  & \textbf{87.94}\scriptsize$\mathbf{\bm\pm 0.18}$  & \textbf{90.52}\scriptsize$\mathbf{\bm\pm 0.13}$  \\ \hline \hline
\end{tabular}}
\end{center}
\vspace{-1.5em}
\end{table}

\subsection{Effectiveness Evaluation}
\label{computational}

Our proposed method targets selecting data for executing off-the-shelf DA operations instead of designing DA operations themselves. Our work is actually DA methods agnostic, in the sense that ours is compatible with any specific DA operations. Thus, we demonstrate the effectiveness of our proposed method based on some of the most representative and commonly used DA approaches including Mixup \cite{zhang2017mixup}, Cutmix \cite{yun2019cutmix} and AutoAugment \cite{cubuk2018autoaugment}.
We follow the common practice in the field of DA method studies \cite{cubuk2018autoaugment,zoph2018learning,gastaldi2017shake} to build the \textit{baseline} model. In \textit{baseline}, we perform very basic data augmentation operations including horizontally flipping with 0.5 probability, zero-padding and random cropping. The data sampling strategies for comparison are: (1) \textit{all}: we perform fully-augmentation in the sense that all images are augmented. (2) \textit{random}: we execute specified DA operations for each sample with a probability $p$ sampled from a uniform distribution $U(0,1)$. (3) \textit{fixed}: we randomly augment each sample with a fixed probability. Here, for different DA strategies, we empirically try different values for the fixed probability and report the one performs the best. As a result, we set $p=0.5$ for CutMix while $p=0.8$ for Mixup, AutoAugment. (4) \textit{linearly}: we linearly increase the probability for executing the corresponding DA operations from 0 to 1 as the training goes by. (5) \textit{onlineRL}: we learn an augmentation probability for each sample independently.


\subsubsection{Classification Results on CIFAR}
The results on CIFAR-10 and CIFAR-100 are reported in Table \ref{table:CIFAR}, which shows that our proposed SelectAugment delivers consistent improvements in the classification accuracy when applied to different DA methods and compared with other data selection strategies. We observe that the SelectAugment is a very general data selection tool that can be applicable for different network architectures and different off-the-shelf DA methods. In the following, we further demonstrate this by applying it on deeper networks and larger datasets.




\subsubsection{Classification Results on ImageNet} 
On ImageNet, as shown in Table \ref{table:imagenet}, our proposed SelectAugment still brings significant improvements in Top-1 accuracy when using ResNet-50 backbone and ResNet-200 backbone respectively. This further demonstrates the effectiveness of our proposed SelectAugment and shows the consistent benefits for the larger dataset and deeper networks.


\subsubsection{Effectiveness on Fine-grained Classification} 
Additionally, we evaluate our proposed method on fine-grained classification. As reported in Table \ref{table:fine-grained}, our proposed SelectAugment is also effective in giving full play to the role of data augmentations through the learned deterministic data selection policy on fine-grained image classification.



\subsection{Ablation Study}

\begin{table}[]
\vspace{0.2em}
\setlength{\abovecaptionskip}{0.3cm}
\begin{center}
\captionsetup{width=1.0\linewidth}
\caption{\textbf{Performance of ablation studies.} We report test accuracy ($\%$) on CIFAR with WRN-28-10 and Top-5 validation accuracy ($\%$) on ImageNet with ResNet-50. We take Mixup as an example for ablation experiments. }
\label{table:ablation}
\scalebox{0.71}{
\begin{tabular}{lcc|ccc}
\hline
\hline
\multicolumn{1}{c}{\multirow{2}{*}{Method}} & \multicolumn{2}{c|}{Policy}   &  \multicolumn{3}{c}{Dataset}             \\ \cline{2-3} \cline{4-6} 
\multicolumn{1}{c}{}                        & Parent & \multicolumn{1}{c|}{Child} &                                                                      CIFAR-10 & CIFAR-100 & ImageNet   \\ \hline
\multicolumn{1}{c}{Only-Parent}                                  & \CheckmarkBold          & \XSolidBrush                                                                      & 97.05    &  82.97    &  77.15   \\
\multicolumn{1}{c}{Only-Child }                                  & \XSolidBrush          & \CheckmarkBold                                                                       & 97.21    & 83.18     & 77.34     \\
\multicolumn{1}{c}{\textbf{SelectAugment}}                                         & \CheckmarkBold          & \CheckmarkBold                                                                    & \textbf{97.33}    & \textbf{83.37}     & \textbf{77.84}    
   \\ \hline \hline
\end{tabular}}
\end{center}
\vspace{-1.2em}
\end{table}

Here, we conduct an ablation study on the hierarchical design of our proposed SelectAugment. Specifically, we compare the SelectAugment to \textit{only-parent} policy learning and \textit{only-child} policy learning, respectively. Their experimental configuration is introduced as below.

\textit{Only-parent} means that a one-level RL algorithm is used to choose the augmentation ratio and samples are randomly selected in the batch under this ratio.

\textit{Only-child} represents that a one-level RL algorithm is adopted to directly decide whether an image is augmented or not in the current batch.



As reported in Table \ref{table:ablation}, our proposed SelectAugment achieves consistent improvements compared to \textit{only-parent} and \textit{only-child} policies. This indicates that the "divide and conquer" idea is more purposeful and effective in selecting data for DA than non-hierarchical designs. The \textit{only-parent} policy learning lacks fine-grained data selection at the instance level. The \textit{only-child} policy learning has a large action space rendering difficult policy exploration.

\vspace{-0.2em}
\subsection{Complexity Analysis}
We analyze the parameters and the time complexity of our proposed SelectAugment. Following the related works in this field \cite{cubuk2018autoaugment,lim2019fast,zhang2019adversarial}, we report the complexity measure results on CIFAR-10 (See Table \ref{table:cost}).   
The parent and child policy network have 0.01M and 1.98M parameters respectively, leading to 1.99M parameter increase in total (less than 6\% of the target network, \ieno, WRN-28-10 with 36.5M parameters). As for the time consuming, our proposed approach will bring 1.4 GPU hours increase in the training time.



\begin{figure}[t]
\vspace{-0.5em}
	\setlength{\abovecaptionskip}{0pt} 
\setlength{\belowcaptionskip}{-0pt}
	\begin{center}
		\includegraphics[width=0.8\linewidth]{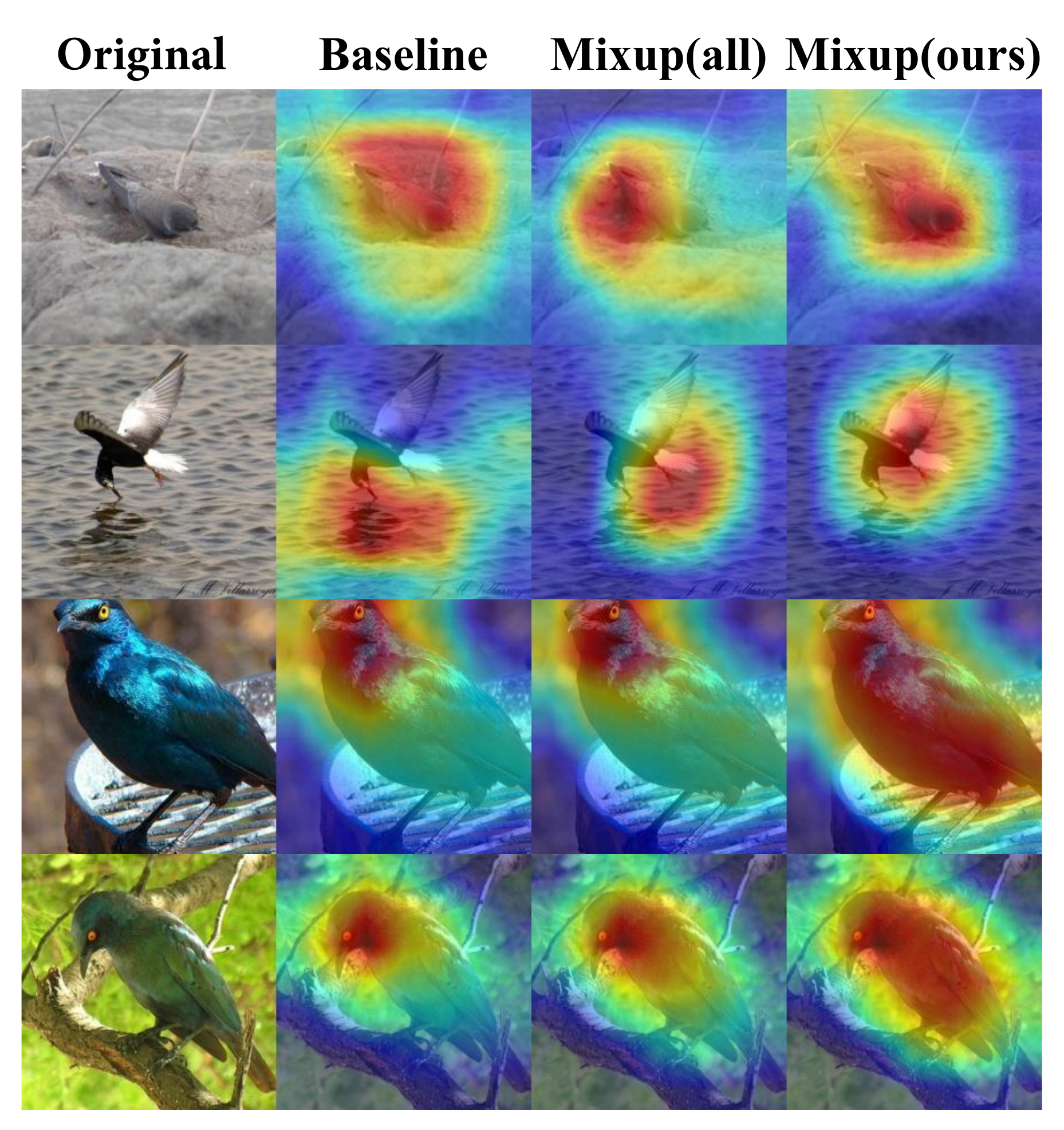}
	\end{center}
	\caption{Grad-CAM \cite{selvaraju2017grad} visualization on the examples from CUB-200-2011 with the models trained by different DA policies. The first column shows original images. The remaining columns show the visualization results for the ResNet-50 trained with the baseline methods, Mixup (all) with fully-augmentation and Mixup (ours) with our proposed SelectAugment, respectively.}
	\label{fig:gradcam} 
\vspace{-1.2em}
\end{figure}

\begin{table}[h]
\setlength{\abovecaptionskip}{0.3cm}
  \begin{center}
\captionsetup{width=1.0\linewidth}
\caption{The architecture parameters, performance and training time comparison of various policies using Mixup on CIFAR-10 dataset with WRN-28-10.}
\label{table:cost}
\scalebox{0.71}{
\begin{tabular}{lccc}
\hline \hline
 Method        & Parameters & Accuracy ($\%$) & GPU hours \\ \hline
\multicolumn{1}{c}{all} &  0M          &    97.14         &      6.5     \\
\multicolumn{1}{c}{Ours}     &  1.99M          &     97.33        &   7.9        \\ \hline \hline
\end{tabular}}
\vspace{-10pt}
\end{center}
\vspace{-0.2em}
\end{table}

\begin{figure}[t]
	\setlength{\abovecaptionskip}{0pt} 
\setlength{\belowcaptionskip}{-0pt}
	\begin{center}
		\includegraphics[width=1.0\linewidth]{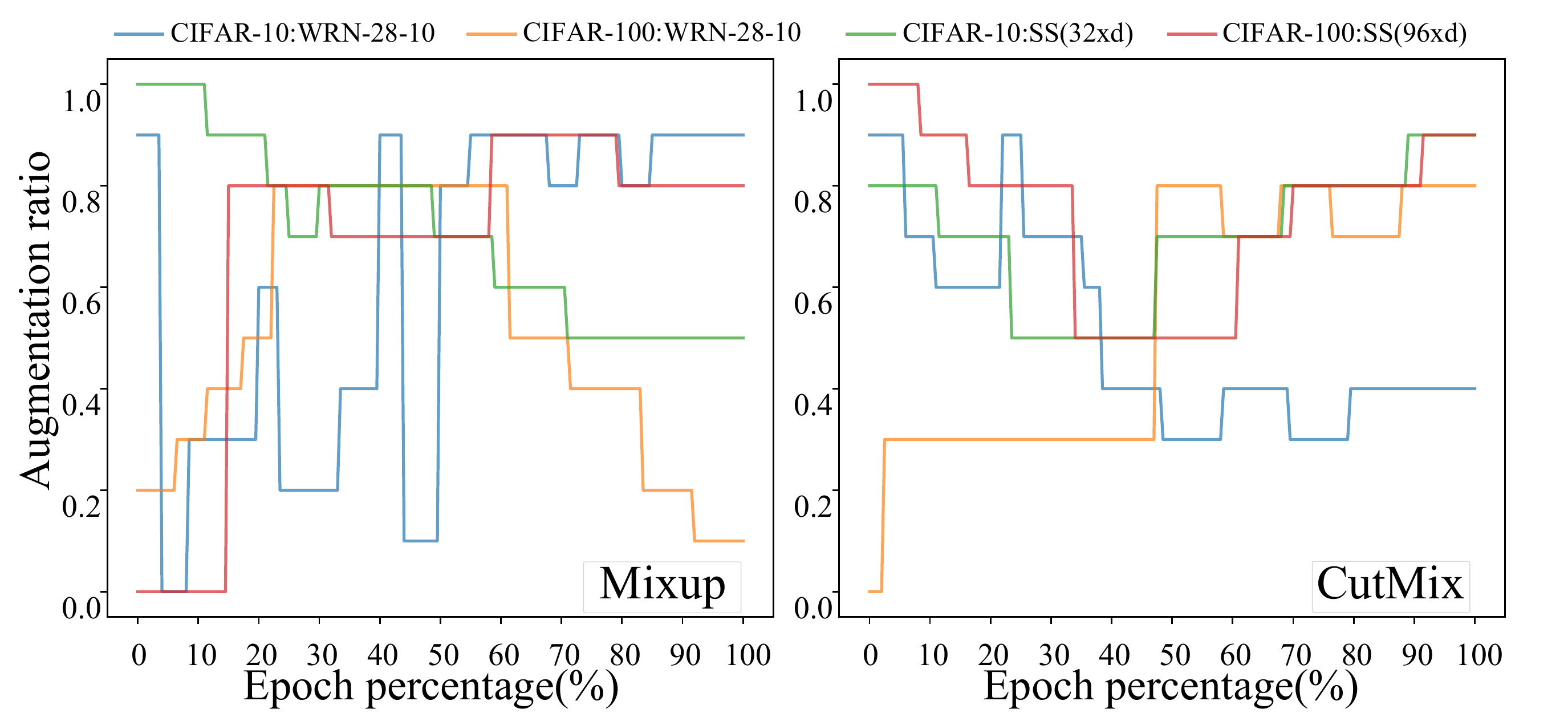}
	\end{center}
	\caption{Visualization augmentation ratio given by parent policy using Mixup and CutMix on CIFAR and various networks. It exhibits that parent policy is not static or a fixed pattern over time. In addition, the policy is different due to changes in datasets, networks and DA methods.  }
	\label{fig:policy} 
\vspace{-1.0em}
\end{figure}


\vspace{-0.2em}
\subsection{Visualization Analysis}

\subsubsection{Grad-CAM Visualization}
Grad-CAM \cite{selvaraju2017grad} is a widely used technique to localize the most important features for classification in a convolutional neural network. We adopt the Grad-CAM to visualize the learned features to study the impact of the data selection by our proposed SeletAugment. As exampled in Fig. \ref{fig:gradcam}, we perform Grad-CAM visualization for 1) Baseline, 2) Mixup (all) with fully-augmentation and 3) Mixup (ours) with SelectAugment which all take the ResNet-50 as the backbone and are trained on CUB-200-2011. Please find more visualization results in the supplementary.



We find that after applying SelectAugment to the DA process for training, the target model (for mainstream task) latches on discriminative cues more precisely and comprehensively. As shown in the first and the second rows of Fig. \ref{fig:gradcam}, the visualization results of the target model trained with SelectAugment indicate that it localizes the foreground region with higher spatial precisions than others. Moreover, the third and fourth rows manifest that the model with SelectAugment localizes more class-related cues.

\subsubsection{Visualization of the Learned Augmentation Ratio} 
We further visualize the parent policy learning results, \ieno, the learned augmentation ratio at the batch level, to analyze the performance of SelectAugment. As shown in Fig. \ref{fig:policy}, we observe that 1) the learned augmentation ratio changes adaptively over different training stages; 2) the learned policy varies a lot for different datasets and network architectures.

\section{Conclusion}

In this paper, we point out that randomly selecting samples to do data augmentation may cause content destruction and visual ambiguities, leading to the distribution shift and thus negatively affecting the target model training. To tackle this, we propose a hierarchical deterministic sample selection strategy to give full play to data augmentation. We adopt HRL to facilitate policy learning towards this goal. Our proposed approach is easy to use and generally applicable for different existing DA methods. Extensive experiments and visualization results demonstrate its effectiveness and versatility for different DA methods, datasets and models.
%


\bibliography{AAAIHDDA}

\newpage

\section{\LARGE {Supplementary Materia}}

\section{Experiment details}

\subsection{Datasets Overview}

For the sake of consistency with the previous works \cite{cubuk2018autoaugment,zhang2017mixup,lim2019fast,yun2019cutmix}, we evaluate SelectAugment on the following datasets: CIFAR-10 \cite{krizhevsky2009learning},  CIFAR-100 \cite{krizhevsky2009learning}, ImageNet \cite{deng2009imagenet} and two fine-grained object recognition datasets (CUB-200-2011 \cite{wah2011caltech}, Stanford Dogs \cite{khosla2011novel}). 

Next, we introduce the basic information of datasets stated above: both CIFAR-10 and CIFAR-100 have 50,000 training examples. Each image of size $32 \times 32$ belongs to one of 10 categories. ImageNet dataset has about 1.2 million training images and 50,000 validation images with 1000 classes. The image size in ImageNet is $224 \times 224$. We evaluate the performance of our proposed SelectAugment on two standard fine-grained classification datasets. CUB-200-2011 \cite{wah2011caltech} consists of 6,000 train and 5,800 test bird images distributed in 200 categories. Stanford Dogs \cite{khosla2011novel} consists of 20,500 images of 120 breeds of dogs. The image size on the two fine-grained datasets is $224 \times 224$.

\subsection{Data Augmentation Methods.} 

Specifically, \textit{Mixup} mixes two samples by interpolating both the images and labels. \textit{CutMix} randomly crops a patch from one image and pastes it to another image. The labels are also mixed proportionally to the number of pixels of combined images. \textit{AutoAugment} contains offline augmentation policies which has information about operation, the probability of using the operation and magnitude. When training the target network, one of these offline policies is randomly selected and performed on a sample. The \textit{baseline} follows previous works \cite{cubuk2018autoaugment,zoph2018learning,gastaldi2017shake}, including horizontally flipping with 0.5 probability, zero-padding and random cropping with 32$\times$32 for CIFAR and 224$\times$224 for ImageNet and CUB-200-2011 and Stanford Dogs.

\begin{figure}[]
	\setlength{\abovecaptionskip}{0pt} 
\setlength{\belowcaptionskip}{-0pt}
	\begin{center}
		\includegraphics[width=1.0\linewidth]{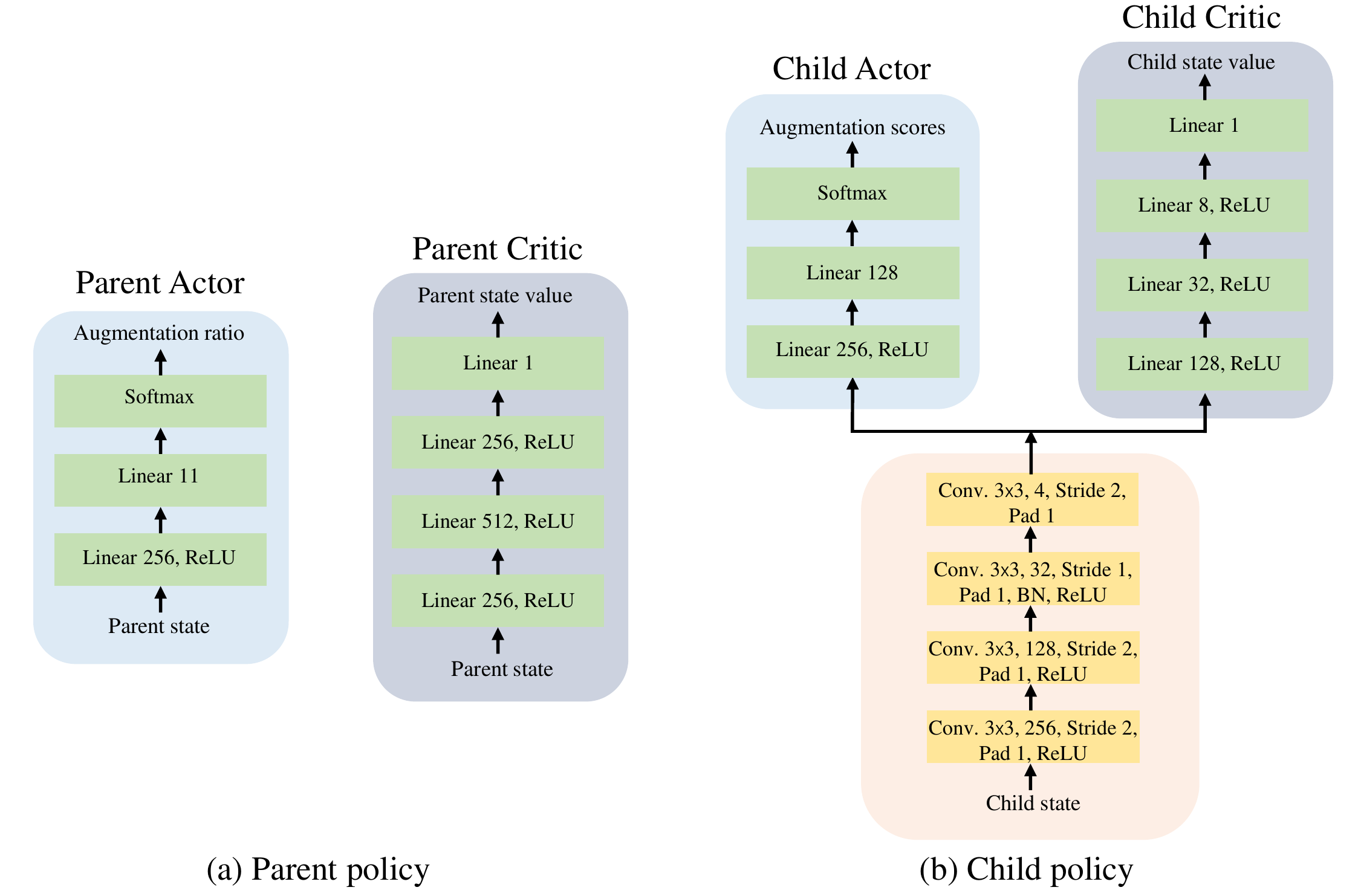}
	\end{center}
	\caption{Our network details including parent actor, parent critic, child actor and child critic.}
	\label{fig:model}
\vspace{-1.5em}
\end{figure}

\subsection{Network Architectures}
In SelectAugment, the parent policy owns a parent actor network and a parent critic network. The parent actor network takes the parent state vector as input to infer a parent action (\ieno augmentation ratio), while the parent critic network appraises the obtained parent state value. Similarly, the child policy also retains a child actor network and a child critic network. Here, the two networks take the child state as input, and output the child action (\ieno, the scores of images suitable for augmentation in a batch) and the child state value respectively. In this section, we provide the detailed model architecture for each network mentioned above in our SelectAugment model, as shown in Fig. \ref{fig:model}.

\subsection{Hyperparameters Details}
For the target model, we detail the target model hyperparameters on CIFAR-10, CIFAR-100, ImageNet, CUB-200-2011 and Stanford Dogs in Table \ref{table:hyperparameters}. We do not specifically tune these hyperparameters, and all of these are consistent with previous works \cite{cubuk2018autoaugment,lim2019fast,zhang2019adversarial,du2020fine,chen2019destruction}. 

In our SelectAugment policy model, for parent policy, we use Adam optimizer with an initial learning rate $1e-5$. For the child policy, we use Adam optimizer with an initial learning rate of $1e-3$, weight decay of $5e-4$ and a cosine learning decay with one annealing cycle. Following the previous works \cite{zhang2017mixup,cubuk2020randaugment,dabouei2021supermix}, we set the batch size $b$ to $128$ for CIFAR and $256$ for Imagenet together with CUB-200-2011 and Stanford Dogs.

\subsection{Training scheme}


In our method, we first use ResNet-18 as a classifier to pre-train the policy network in SelectAugment. This is because that the cold start problem \cite{wei2016collaborative} exists in RL. That is, in the initial stage of interaction, RL is still exploring the environment, so the actions selected according to the initial policy may not be appropriate, resulting in a poor strategy. The common solution is to find a small similar network to pre-train the policy network, and then continue to use the pre-trained network on the true target network to avoid meaningless actions in the initial stage. Therefore, through using the pre-train policy, our hierarchical policy has certain decision-making capabilities even in the initial stage of training the target network. It improves policy search efficiency and thereby improves the performance of the target mainstream network.


\begin{table}[]
\begin{center}

\caption{Model hyperparameters on CIFAR-10,CIFAR-100, ImageNet, CUB-200-2011 (CUB) and Stanford Dogs (DOG). LR represents learning rate of the target network, WD represents weight decay, and LD represents learning rate decay method. If LD is multistep (multi), we decay the learning rate by 10-fold at epochs 30, 60, 90 etc. according to LR-step. }
\label{table:hyperparameters}
\scalebox{0.68}{
\begin{tabular}{c|c| c c c c |c}
\hline
\hline
Dataset&Model  & LR & WD & LD &LRstep& Epoch \\
\hline
\multirow{2}{*}{CIFAR-10} & WRN  & 0.1 & 5e-4 & cosine &- &200  \\
&SS(26 2x32d) & 0.2 & 1e-4 & cosine&- &600\\

\hline
\multirow{2}{*}{CIFAR-100} &WRN-28-10&0.1&5e-4&cosine&-&200\\
&SS(26 2x96d)  &0.1 &5e-4&cosine &-&1200\\

\hline
\multirow{2}{*}{ImageNet} &ResNet-50&0.1&1e-4&multi&[30,60,90,120,150] &270\\
&ResNet-200  &0.1 &1e-4&multi& [30,60,90,120,150] &270\\
\hline
\multirow{2}{*}{CUB} &ResNet-50&1e-3&1e-4&multi&[30,60,90]&200\\
&ResNet-152  &1e-3 &1e-4&multi&[30,60,90]  &200\\
\hline
\multirow{2}{*}{Dog} &ResNet-50&1e-3&1e-4&multi&[30,60,90]& 200\\
&ResNet-152 &1e-3 &1e-4&multi&[30,60,90] & 200\\
\hline
\hline
\end{tabular}}
\end{center}  

\end{table}

\section{Additional Experiments}

\subsection{An extension \textit{SelectAugment+}}

We introduce SelectAugment+ as a strengthened version of our proposed SelectAugment. SelectAugment+ further learns a policy to select the most suitable DA operations for the samples selected by our proposed SelectAugement. We first detail our design of SelectAugment+. Furthermore, we place the experimental results of SeletAugment+ and propose some analysis.

\textbf{SelectAugment+ Design}
We formulate the task of DA operation selection as a decision-making problem and adopt RL to search for an effective policy. To better understand, we call it the operation policy. With the consideration of samples content and semantics, the operation policy chooses a relatively appropriate DA operation for samples from our pre-defined DA methods pool. We design the MDP tuple of the operation policy in RL in detail. 

\textit{State} Similar with the child policy in SelectAugment, the operation policy also needs to perceive the sample contents. Therefore, we utilize the deep features of images extracted by target network as the state of this policy.

\textit{Action} The operation policy chooses a DA operation for each sample in a batch from the DA methods pool (\ieno, the action space of the operation agent). We define the operation pool including Mixup \cite{zhang2017mixup},CutMix \cite{yun2019cutmix}, Cutout \cite{devries2017improved} and AutoAugment \cite{cubuk2018autoaugment}. In other words, the DA operation is chosen for each sample from commonly used DA methods with superior performance. Moreover, the image processing operations do not overlap between these methods. Note that if AutoAugment is chosen as action, one of the offline policies is randomly selected to be applied to the sample. Here, we define the chosen DA operations for samples in the same batch as a vector $\bm{\psi(\cdot)}$ with batch-size dimension. 

We make the samples chosen by SelectAugment to execute augmentation operations selected by the operation policy as follows:
\begin{equation}
  \bm{x_{aug}}\!=\!\{ \psi_i(x_i)|a_i^C \in topK({\mathbf{a^C}}),x{_i} \in \bm{x}, \psi_i(\cdot) \in \bm{\psi(\cdot)}\},
  \label{equ:topK}
\end{equation}
\begin{equation}
  \bm{x_{ori}} =\!\{ x_j|a_j^C \notin topK({\mathbf{a^C}}),x{_j} \in \bm{x}\},
  \label{equ:ori}
\end{equation}
where $\bm{x}$ denotes the original training mini-batch (without any augmentation operations) and $\bm{\widetilde x}=\{ \bm{x_{aug}},\bm{x_{ori}}\}$ represents the selectively augmented mini-batch processed by SelectAugment+.

\textit{Reward Function} The parent policy, the child policy and the operation policy serve for the same goal (\ieno, enhancing the performance of the target network), therefore, we use the same reward function design for the three policies in SelectAugment+. The design of reward function is similar to reward design in SelectAugment, except that $\bm{\widetilde x}$ represents the augmented mini-batch processed by SelectAugment+. The reward function of the three policies is formulated as:
\begin{equation}
{r} =  l(\phi (\bm{x}),\bm{b})-l(\phi (\bm{\widetilde x}),\bm{y}) + l(\phi (\bm{\overline x} ),\bm{y})-l(\phi (\bm{\widetilde x}),\bm{y}). 
\label{equ:reward}
\end{equation}

\begin{figure*}[]
	\setlength{\abovecaptionskip}{0pt} 
\setlength{\belowcaptionskip}{-0pt}
	\begin{center}
		\includegraphics[width=0.8\linewidth]{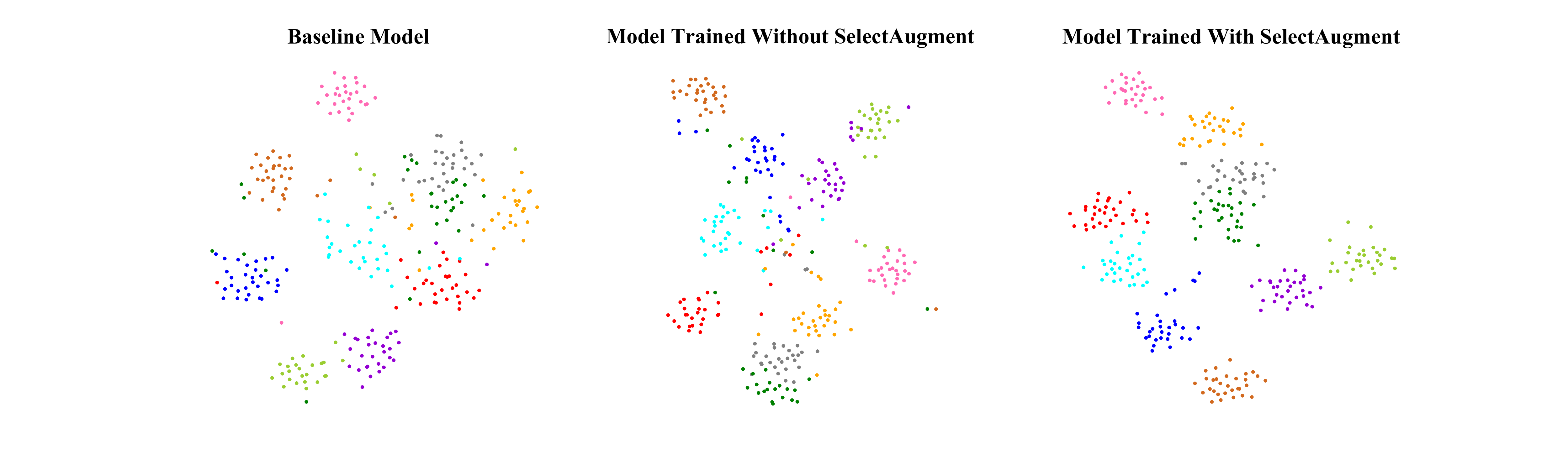}
	\end{center}
	\caption{T-SNE Visualization of feature vectors of test data extracted by models trained with baseline (left), AutoAugment which augments all images (middle) and AutoAugment with SelectAugment (right). }
	\label{fig:tsnemodel}
\vspace{-1em}
\end{figure*}

\begin{figure*}[]
	\setlength{\abovecaptionskip}{0pt} 
\setlength{\belowcaptionskip}{-0pt}
	\begin{center}
		\includegraphics[width=0.8\linewidth]{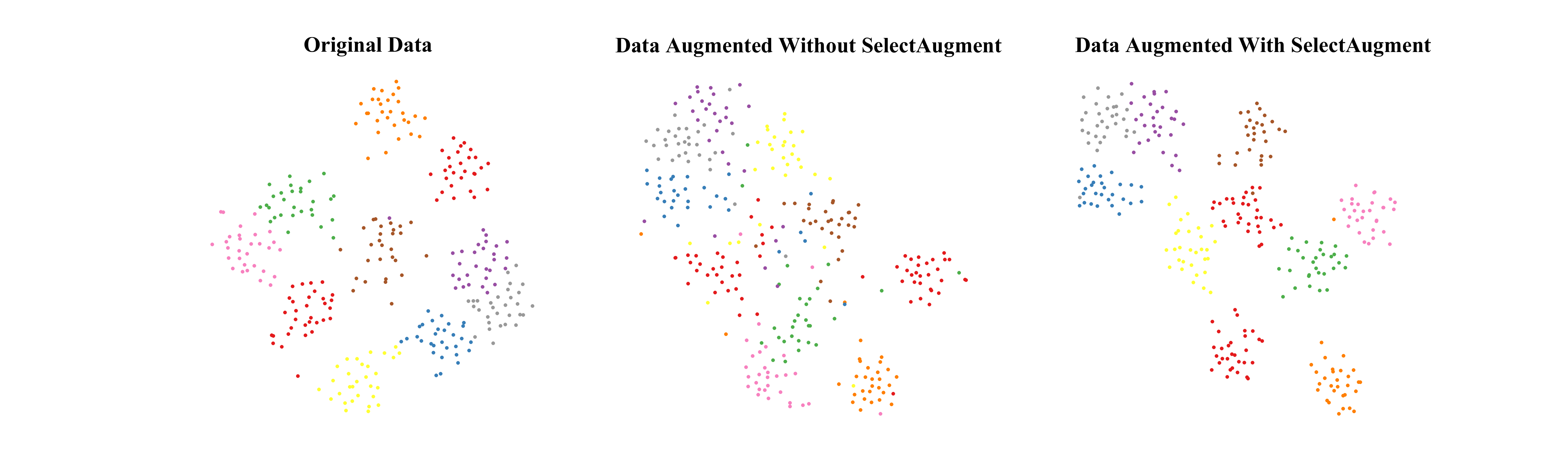}
	\end{center}
	\caption{T-SNE Visualization of feature vectors of different processed data (\ieno, original data (left), augmented data using AutoAugment without SelectAugment (middle) and augmented data using AutoAugment with SelectAugment (right)) extracted by the pre-trained ResNet-50.}
	\label{fig:tsnedata}
\vspace{-1em}
\end{figure*}

\textit{Policy Learning} The operation policy and the two policies (\ieno, the parent and child policy) are learned simultaneously and together with the target network. The parent and child strategy remain the same as described in the body text. We adopt A2C algorithm to learn the operation policy due to the discrete control in the operation policy. Therefore, the operation policy model also owns a actor network and a critic network whose architecture is similar with the child policy, except the last layer is modified to a \textit{Softmax}. Therefore, the loss functions of updating these two networks is consistent with the loss functions of child policy. Please refer to Eq.(8, 9, 10) in the body text.


\textbf{SelectAugment+ Results on CIFAR}
The SelectAugment+ results on CIFAR-10 and CIFAR-100 are reported in Table \ref{table:augmentplus}. Concretely, SelectAugment+ which selects an appropriate DA method for each SelectAugment-selected sample outperforms SelectAugment combined with one DA method. It demonstrates that taking operation selection into consideration will further enhance the effect of DA. Moreover, this reveals that DA operation is closely related to the content of the image rather than treating them indiscriminately using the same DA method. 

\subsection{More Ablation Experiments}

In the body text, we compare our hierarchical policy against one-level policy (\ieno, \textit{only-parent} using RL to choose ratio and \textit{only-child} using RL to directly decide which images to augment in a batch) and take Mixup \cite{zhang2017mixup} as an example. Here, we conduct more ablation experiments using AutoAugment \cite{cubuk2018autoaugment} to cogently confirm the advantage of two-level policies as shown in Table \ref{table:ablationaa}.

\subsection{Discussion of the Design in SelectAugment}
In this section, we discuss the important designs (\ieno the ratio pool) in HRL framework in SelectAugment.

\textbf{Ratio Pool}
The parent policy selects an appropriate ratio from the ratio pool as the augmentation ratio for the batch. Considering the automatic policy aims to relieve human expertise, we select discrete values at equal intervals from $[0,1]$ to form the ratio pool. Here, we set the interval number to $10, 20, 40, 100$ respectively to discuss the impact of the configuration of the augmentation ratio pool. As shown in Table \ref{table:pool}, regardless of the interval number, the performance of 
our method has a consistent improvement compared with the random selection policy \textit{all}. In addition, we observe that the interval number $10$ (\ieno, the ratio pool $[0,0.1,0.2,...,1.0]$) performs the best on different datasets and networks.
Actually, the interval number determines the granularity of actions and the size of action space.
Specifically, the smaller the interval number is, the more coarse the parent control policies are.
Extremely, setting the interval number to 1 means that we can only choose to augment all samples or keep all samples original. Extreme large interval number allows us to have more fine-grained controls but renders a significant increase in the size of action space, which affects the difficulties of policy exploration. Therefore, there is a trade-off between the control granularity and the optimization difficulties corresponding to the size of the action space.

\begin{table}[H]
\vspace{-0.5em}
\setlength{\abovecaptionskip}{0.3cm}
\begin{center}
\captionsetup{width=1.0\linewidth}
\caption{\textbf{Performance of different augmentation ratio pool together with different batch size configurations.} We report test accuracy ($\%$) on CIFAR with WRN-28-10. We take Mixup as an example. }
\label{table:pool}
\label{table:pool}
\scalebox{0.71}{
\begin{tabular}{c|c|c|ccccc}
\hline \hline
 \multicolumn{1}{c|}{\multirow{2}{*}{\begin{tabular}[c]{@{}c@{}}Dataset $\&$\\ Model $\&$ DA\end{tabular}}} & \multirow{2}{*}{Batch Size} & Policy & \multicolumn{5}{c}{Interval Number}    \\ \cline{3-8} 
                         &                        & all    & 1             & 5    & 10 &20   & 100   \\ \hline
\multirow{4}{*}{\begin{tabular}[c]{@{}c@{}}CIFAR-100\\ WRN-28-10\\ Mixup\end{tabular}}                & 32              & 82.28  & 82.05 & 82.64 &  \textbf{83.22} &83.10  & 83.06 \\
           & 64             & 82.36  & 82.11 & 82.68 & \textbf{83.25} & 83.17 & 83.12 \\
                           & 128            & 82.41 & 82.15 & 82.82 & \textbf{83.37} & 83.21 & 83.15 \\
                                                & 256            & 82.32  & 82.17 & 82.79 & \textbf{83.32} & 83.19 & 80.72 \\

                      \hline \hline
\end{tabular}}
\end{center}
\vspace{-0.6em}
\end{table}

\textbf{Batch Size}
We provide detailed comparison results on different batch sizes as shown in Table \ref{table:pool}. It shows that our proposed SelectAugment delivers consistent gains. Our proposed method is not sensitive to the choice of batch size.


\begin{table}[]
\vspace{0.2em}
\setlength{\abovecaptionskip}{0.3cm}
\begin{center}
\captionsetup{width=1.0\linewidth}
\caption{\textbf{Performance of ablation studies.} We report test accuracy ($\%$) on CIFAR with WRN-28-10 and Top-5 validation accuracy ($\%$) on ImageNet with ResNet-50. We take AutoAugment as an example for ablation experiments. }
\label{table:ablationaa}
\scalebox{0.71}{
\begin{tabular}{lcc|ccc}
\hline
\hline
\multicolumn{1}{c}{\multirow{2}{*}{Method}} & \multicolumn{2}{c|}{Policy}   &  \multicolumn{3}{c}{Dataset}             \\ \cline{2-3} \cline{4-6} 
\multicolumn{1}{c}{}                        & Parent & \multicolumn{1}{c|}{Child} &                                                                      CIFAR-10 & CIFAR-100 & ImageNet   \\ \hline
\multicolumn{1}{c}{Only-Parent}                                  & \CheckmarkBold          & \XSolidBrush                                                                      &   97.35  &  83.22    &  77.30   \\
\multicolumn{1}{c}{Only-Child }                                  & \XSolidBrush          & \CheckmarkBold                                                                       &  97.48   & 83.52     & 77.81     \\
\multicolumn{1}{c}{\textbf{SelectAugment}}                                         & \CheckmarkBold          & \CheckmarkBold                                                                    & \textbf{97.65}    & \textbf{83.81}     & \textbf{78.16}    
   \\ \hline \hline
\end{tabular}}
\end{center}
\vspace{-1.2em}
\end{table}

\begin{table*}[h]
\caption{The performance of SelectAugment+. We report test accuracy ($\%$) on CIFAR-10 with WRN-28-10 and SS(26 2x32d), CIFAR-100 with WRN-28-10 and SS(26 2x96d). Mixup(all) and AutoAug(all) represents that all samples are augmented by Mixup and AutoAugment, respectively. Mixup(ours) and AutoAug(ours) represents that samples selected by SelectAugment to be augmented by Mixup and AutoAugment, respectively. }
\label{table:augmentplus} 
\begin{center}
\scalebox{0.71}{
\begin{tabular}{c|c|cccccc}
\hline
\hline
\multicolumn{1}{c|}{\multirow{2}{*}{Dataset}} & \multicolumn{1}{c|}{\multirow{2}{*}{Model}} & \multicolumn{6}{c}{Policy}              \\ \cline{3-8} 
\multicolumn{1}{c|}{}                         & \multicolumn{1}{c|}{}                       & Baseline & Mixup(all) & Mixup(ours) & AutoAug(all) & AutoAug(ours) & SelectAugment+ \\ \hline
CIFAR-10                                      & WRN-28-10                                          &  96.13        &   97.14    & 97.33&97.28&97.65&\textbf{97.72}          \\
CIFAR-10                                      & SS(26 2x32d)                                          &  96.26        &   97.12    & 97.38&97.37&97.61&\textbf{97.68}       \\
CIFAR-100                                      & WRN-28-10                                        &  81.20        &   82.41   & 83.37 &83.08&83.81&\textbf{84.11}        \\
CIFAR-100                                     &  SS(26 2x96d)                        &  82.85        &   84.77    & 85.17 &85.66&85.87&\textbf{85.94}     \\ \hline
\hline
\end{tabular}}
\end{center}

\end{table*}

\section{Additional Visualization}





\subsection{More Grad-CAM Visualization}
\begin{figure*}[]
	\setlength{\abovecaptionskip}{0pt} 
\setlength{\belowcaptionskip}{-0pt}
	\begin{center}
		\includegraphics[width=0.9\linewidth]{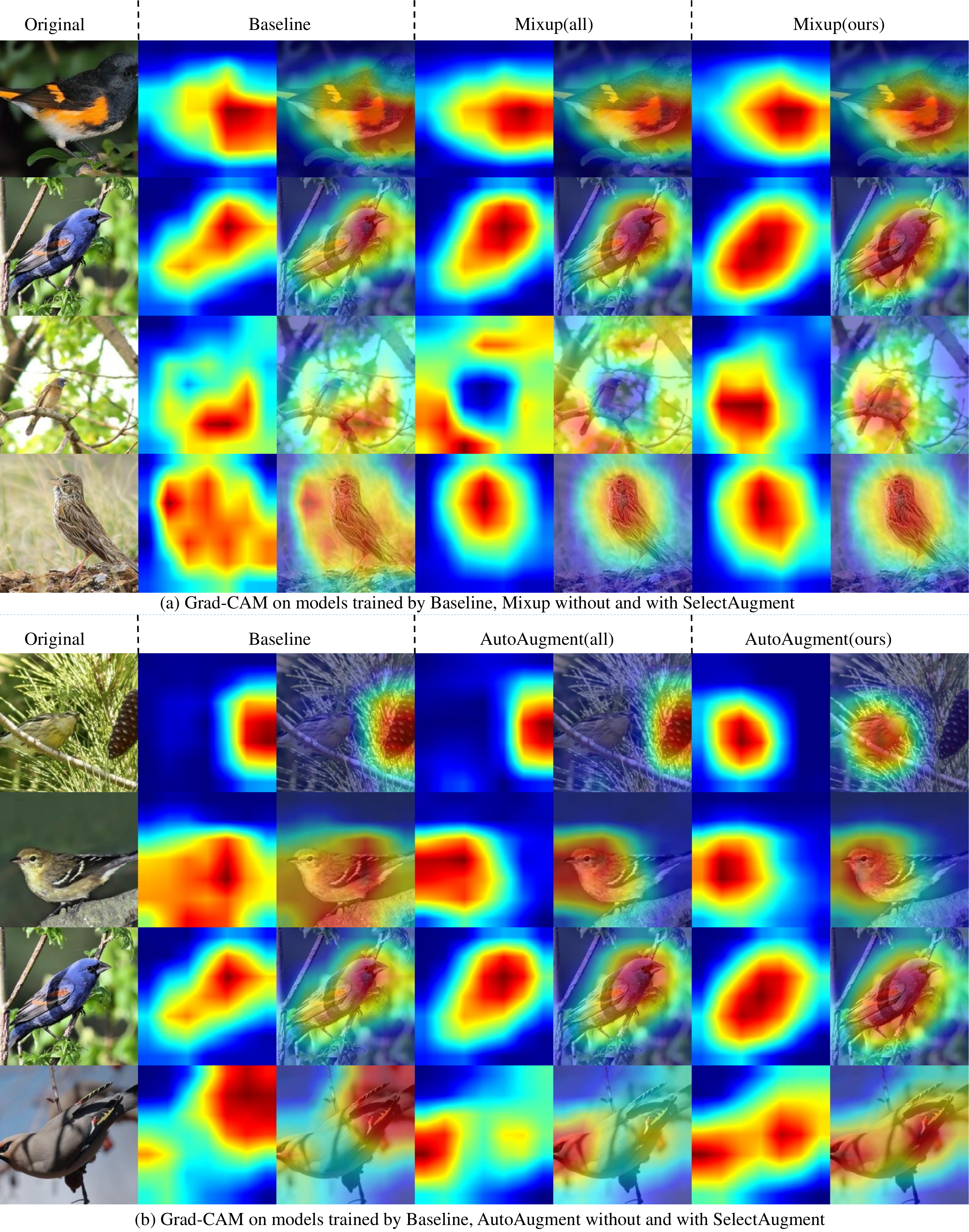}
	\end{center}
	\caption{ The visualization of features (left column is heatmap and right column is heatmap mixed with original images) extracted by ResNet-50 trained with baseline model, \textit{all} and SelectAugment using Grad-Cam\cite{selvaraju2017grad}. For images with complex backgrounds, SelectAugment can better help target network to locate foreground objects. Meanwhile, target network focus on more parts of the foreground object with the help of SelectAugment. In short, SelectAugment tends to help the target network focus on more discriminative areas. }
	\label{fig:gradcammore}
\vspace{-1.5em}
\end{figure*}

To verify that SelectAugment is indeed impacted effectively on the target network, we visualize the regions in which a CNN focuses much attention using Grad-CAM \cite{selvaraju2017grad}. In this section, we provide more Grad-CAM results of ResNet-50 models trained using baseline augmented data, Mixup with fully-augmentation images and Mixup with SelectAugment, respectively, as shown in Fig. \ref{fig:gradcammore}~ (a). Furthermore, we replace Mixup with AutoAugment and then show Grad-CAM visualization results of ResNet-50 models trained using baseline, AutoAugment with \textit{all} which fully augments data and AutoAugment with processed data by our proposed SelectAugment respectively, as illustrated in the Fig. \ref{fig:gradcammore} (b).

The Grad-CAM results demonstrate that our proposed SelectAugment improves the localization ability of the target network and tends to help the target network focus on more parts of the foreground object. In short, our proposed SelectAugment can make the target network focus on the important or representative regions (\ieno, the foreground object closely related to the given label) in the image. As discussed before, uncontrollable executing DA operations without explicitly taking into account the samples may incur unrecoverable information loss and thus optimization biases. SelectAugment provides more accurate clues to the target mainstream network, as \cite{srivastava2014dropout,choe2019attention,uddin2020saliencymix} claimed, and thereby alleviates the optimization bias and improves the performance of the target network.


\subsection{Distribution Visualization and Analysis}
The t-SNE \cite{van2008visualizing} method is adopted to represent high-dimensional feature vectors of data extracted by a model in a low-dimensional space to visualize it.  
We analyze the feature distribution from two aspects, namely the model trained with SelectAugment and the data itself processed by SelectAugment.


First, we respectively use ResNet-50 models trained by baseline, without SelectAugment, and with SelectAugment as feature extractors to display the results of t-SNE on the test data, as demonstrated in Figure \ref{fig:tsnemodel}. In order to make the results intuitive and clear, we only show ten categories. We observe that if the model is trained without a deterministic data selection policy, the features extracted by the model from different categories are entangled together, which is very confusing for the classifier. In contrast, the features extracted by the model trained with SelectAugment have more obvious boundaries and less mess, which is conducive to the improvement of task network performance. 

Besides, inspired by \cite{yang2021free,xue2019selective,ye2020synthetic}, we use pre-trained ResNet-50 model as the feature extractor to 
conduct t-SNE on the original dataset, augmented datasets using AutoAugment without SelectAugment and with SelectAugment (see Figure \ref{fig:tsnedata}).
The middle row denotes that some samples augmented by AutoAugment without SelectAugment are far away from the class. The reason may be that random selection might lead to negative samples, which miss important and representative features while are far away from the class centre that contains all the representative features, as the "broken" examples shown in Fig. 1 in the main text. 

Moreover, we calculate the distance of each sample to the class centroid and form a histogram to simulate the feature distribution of the dataset, as shown in Fig. 2 in the main text. Obviously, the distribution shift exists between the augmented dataset without selection and the original dataset. As stated before, the feature distribution bias leads to side effects on the target network. On the contrary, our proposed SelectAugment explicitly determines whether to augment an image based on its content and the status of target network to alleviate the feature distribution bias. That is, images are augmented without lost and damaged representative features, which can improve the performance and generalization of target network. 





\subsection{More Discussion on toy examples} The Fig. \ref{fig:negative} illustrates two simple examples of content destruction and visual ambiguities. Actually, the data selection problem for executing DA operations is a complicated one because its optimal policy depends on various factors including the dynamic training status of neural networks as well as the contents of samples. This is why it is quite difficult to ascertain how likely the side-effects will happen and what transformation will produce side-effects, as the learning status varies over time. This is also the reason we illustrate the side-effects of DA attributable to its randomness by visualizing the caused distribution shift (see Fig. \ref{fig:distribution}) from the perspective of statistics. Based on the delicate consideration above, we thus propose to adopt a hierarchical RL based method to solve this problem. Besides, worthy of mention is that the side-effects are not limited to being caused by random cropping operations. To name a few, 1) applying rotation on number $``6”$ may cause confusion with $``9”$; 2) applying DA on hard original images which causes excessive augmentation damaging the training; 3) applying translation on the image with the object at the edge which causes object removed, \etcno.


\end{document}